\documentclass[acmlarge]{acmart}
\usepackage{color}
\usepackage{graphicx}
\usepackage{subfigure}
\usepackage{diagbox}
\usepackage{threeparttable}
\usepackage{fancyhdr}
\usepackage{booktabs}
\usepackage{multirow}
\usepackage{rotating}
\usepackage{pdflscape}
\usepackage[T1]{fontenc}
\usepackage{makecell}
\usepackage[utf8]{inputenc}
\usepackage{fancyhdr}
\AtBeginDocument{%
  }

\setcopyright{acmcopyright}
\copyrightyear{2023}
\acmYear{2023}
\acmDOI{XXXXXXX.XXXXXXX}

\acmPrice{15.00}
\acmISBN{978-1-4503-XXXX-X/18/06}

\begin{document}

\title{Acceleration for Deep Reinforcement Learning using Parallel and Distributed Computing: A Survey}

\author{Zhihong Liu}
\email{zhliu@nudt.edu.cn}
\orcid{0000-0003-3830-7182}
\author{Xin Xu}
\email{xinxu@nudt.edu.cn}
\orcid{0000-0003-3238-745X}
\author{Peng Qiao}
\email{pengqiao@nudt.edu.cn}
\orcid{0000-0001-6752-7892}
\author{Dongsheng Li}
\email{dsli@nudt.edu.cn}
\orcid{0000-0001-9743-2034}

\affiliation{%
  \institution{National University of Defense Technology}
  \streetaddress{109 Deya Rd}
  \city{Changsha}
  \state{Hunan}
  \country{China}
  \postcode{410073}
}

\renewcommand{\shortauthors}{Liu et al.}

\begin{abstract}
 Deep reinforcement learning has led to dramatic breakthroughs in the field of artificial intelligence for the past few years. As the amount of rollout experience data and the size of neural networks for deep reinforcement learning have grown continuously, handling the training process and reducing the time consumption using parallel and distributed computing is becoming an urgent and essential desire. 
 %Therefore, leveraging parallel and distributed computing to tackle this challenge, with techniques ranging from hardware architecture design to parallel algorithms, has become a hot research topic recently. 
 In this paper, we perform a broad and thorough investigation on training acceleration methodologies for deep reinforcement learning based on parallel and distributed computing, providing a comprehensive survey in this field with state-of-the-art methods and pointers to core references. %In particular, we demystify the details from perspectives of key theory, methodologies and techniques in this field, including learning system architectures, simulation parallelism, model synchronization, computing parallelism, deep evolutionary reinforcement learning, and open-source libraries and platforms. Based on these approaches, we extrapolate potential directions for future work.
 In particular, a taxonomy of literature is provided, along with a discussion of emerging topics and open issues. This incorporates learning system architectures, simulation parallelism, computing parallelism, distributed synchronization mechanisms, and deep evolutionary reinforcement learning. 
 Further, we compare 16 current open-source libraries and platforms with criteria of facilitating rapid development. Finally, we extrapolate future directions that deserve further research. 
\end{abstract}





\maketitle

\section{Introduction}

Since the advent of Deep Q Network (DQN)\cite{mnih2015human} in 2015 which achieved human-level control on Atari video games, deep reinforcement learning (DRL), a powerful machine learning paradigm, has been widely investigated by Artificial Intelligence (AI) researchers. DRL deals with training an agent to learn the optimal policy based on feedback from interaction with the environment. Through leveraging the power of deep learning and reinforcement learning, there are many merits in DRL. First, DRL is able to handle high-dimensional and large state spaces. Second, DRL has the general self-learning ability without the requirements of labeled datasets. Third, %DRL can ingeniously transform online computation in traditional solutions to offline training manner. 
problems that need to be solved by online computation traditionally can be solved in offline training manner with DRL. With these merits, DRL in recent years has led to dramatic breakthroughs in game playing\cite{Silver2016,Silver2017,vinyals2019grandmaster}, robotics\cite{kaufmann2023champion, Andrychowicz2020,fan2018surreal,Gupta}, healthcare\cite{popova2018deep,yala2022optimizing,saria2018individualized}, etc.
%\cite{zheng2022ai,}, etc. 

However, with the increased complexity of the applications for DRL, more interaction iterations and larger scale of neural network models are required. As a result, the training process becomes very computation-intensive and thus time-consuming. For instance,  it needs around 38 days of experience to train DQN on Atari 2600 game~\cite{mnih2015human} with 50 million frames. % showed that 7 days is needed to train simulated 3D locomotion tasks, %alphazero 
%AlphaZero required 5000 first-generation TPUs to generate experiences and 64 second-generation TPUs to train the neural network model~\cite{Silver2017},  
Besides, tuning of hyper-parameters is tricky and further increases the time consumption~\cite{henderson2018deep}. 

%3D locomotion tasks

%\begin{figure}[t]
%	\centering
%	\includegraphics[width=0.48\textwidth]{figures/data.png}
%	\caption{Web of Science topic search for "deep reinforcement learning" and (distributed or parallel) (2023.01.19.)".}
%	\label{fig:data}       % Give a unique label
%\end{figure}

%acceleration for large-scale requirement

Thus, it is intuitive to leverage parallel and distributed computing in DRL to accelerate its training process.  During the past few years, rapid progress has been witnessed in this field, resulting in a profusion of studies that target diversified aspects and use a variety of techniques. This causes significant difficulties in understanding the development of this field in a systematic manner. 

In this survey, we collect, classify, and compare a huge body of work on DRL acceleration using parallel and distributed computing, providing a comprehensive survey in this field with state-of-the-art methods and pointers to core references. In particular, we analyze the primary challenges to make DRL training distributed and demystify the details of the technologies that have been proposed by researchers to address these challenges. This includes system architectures for distributed DRL, simulation parallelism, computing parallelism, distributed synchronization mechanisms (\emph{backpropagation-based training}), and deep evolutionary reinforcement learning (\emph{evolution-based training}). 
Furthermore, we provide an overview and comparison of current open-source libraries and platforms that put parallel and distributed DRL into practice. Based on that, we extrapolate potential directions for future work.

\begin{figure*}[t]
	\centering
	\includegraphics[width=0.83\textwidth]{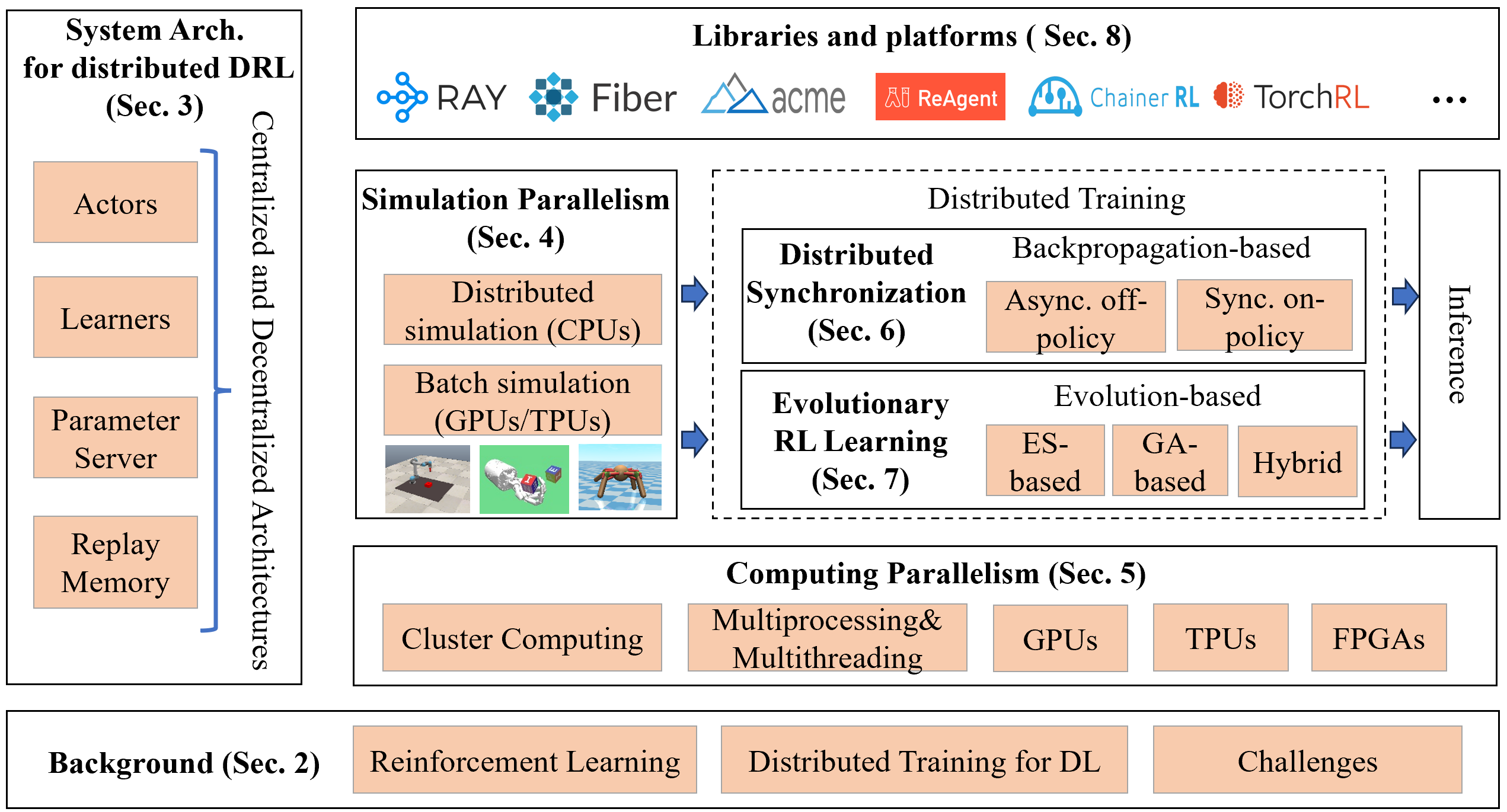}
	\caption{The structure of the survey.}
	\label{fig:stru}       % Give a unique label
\end{figure*}

\subsection{Related surveys}
There are a number of surveys in this field that are related to ours.  Arulkumaran et al.~\cite{Arulkumaran2017} provide a brief survey on DRL algorithms, applications, and challenges. Li~\cite{li2017deep} gives a wide coverage of core elements, important mechanisms, and applications in DRL.  Wang et al.~\cite{9904958} present an overview of the theories, algorithms, and research topics of DRL. Moerland et al. \cite{moerland2023model} provide the taxonomy, key challenges, and benefits of model-based reinforcement learning. Meanwhile, many surveys targeting specific application domains have emerged in recent years, e.g., autonomous driving\cite{Kiran2022}, cyber security\cite{Nguyen2021a}, networking\cite{Luong2019}, intelligence transportation\cite{Haydari2022}, multi-agent systems~\cite{Nguyen2020} and Internet of Things~\cite{chen2021deep}. However, these surveys either focus on the fundamental theories and general technologies of DRL, or investigate DRL methods in specific domains. Few of them studies on learning acceleration by parallel and distributed computing.

Admittedly, there are several surveys have been proposed in distributed machine learning \cite{verbraeken2020survey,peteiro2013survey,wang2020survey}. It is noted that reinforcement learning is one of the sub-branches in machine learning. These surveys, however, summarize distributed technologies from a holistic perspective of machine learning, without in-depth discussion regarding distributed reinforcement learning. Besides, many works have emerged on distributed deep learning training acceleration. Mayer et al.~\cite{Mayer2020}  present a comprehensive investigation on challenges, techniques, and frameworks for DL training on distributed infrastructures. Ben-Nun et al.~\cite{Ben-Nun2019} provide a concurrency analysis for parallel and distributed DL. Tang et al. provide a comprehensive survey of distributed deep learning from the communication perspective. Chahal et al.\cite{Chahal2020} propose a hitchhiker's guide on distributed training for DL. However, these surveys target at DL and have not analyzed any methods in DRL. Note that DL and DRL are different types of machine learning paradigms, and the key difference exactly lays on the training scheme. With respect to DL, the methods learn from supervised training datasets and seek to minimize the error between the neural network model and the labeled data. In comparison, DRL methods require interaction with environments to generate experiences. Based on that, DRL methods learn what actions lead to the maximum outcome. This will unavoidably cause differences in training acceleration techniques between DL and DRL. %, e.g., learning system architecture design, sample generation parallelism, model synchronization, and computing parallelism. 
%Therefore, copying the distributed training acceleration methods in DL to DRL is infeasible.

In particular, Samsami et al.~\cite{Samsami2020} provide a brief overview of distributed DRL by introducing several key research works in this field. No taxonomy of existing methods is presented in their work. Besides, Yin et al. \cite{yin2024distributed} propose a survey on distributed DRL for the specific field of multi-player multi-agent learning, where agents are either cooperating or competing within the environments. In contrast, our survey takes a much more comprehensive and in-depth view on training acceleration by parallel and distributed computing. Along with the outlined taxonomy and literature overview, the details of key methodologies in DRL training are demystified. %Besides, we also point out the future directions that deserve further research. 
To the best of our knowledge, this is the first comprehensive survey on acceleration for DRL using parallel and distributed computing.

\subsection{Structure of the survey}
The structure of the survey is summarized in Fig. \ref{fig:stru} and organized as follows. 

\begin{itemize}
	\item Section \ref{sec:bg} provides the foundations and challenges related to our study.
	\item Section \ref{sec:arch} classifies system architectures in general for parallel and distributed DRL. 
	\item Section \ref{sec:sim} elaborates on different simulation parallelism strategies to expose simulation concurrency.
	\item Section~\ref{sec:compu} analyzes computing parallelism patterns used in existing works.
	\item Section~\ref{sec:sync} discusses distributed synchronization mechanisms for backpropagation-based distributed training methods.
	\item Section~\ref{sec:ec} explores technologies of deep evolutionary reinforcement learning for evolution-based distributed training methods.
	\item Section~\ref{sec:lib} compares current open-source libraries and platforms that put parallel and distributed DRL into practice. 
	\item Section~\ref{sec:conclusion} gives concluding remarks and highlights future directions that deserve further research.  
\end{itemize}

\section{Background}\label{sec:bg}
Deep reinforcement learning (DRL) integrates deep learning (DL) and reinforcement learning (RL), where DL is leveraged to perform function approximation in RL. Thus, the basic knowledge of DRL and current distributed DL training development lay a foundation for parallel and distributed DRL. This section will discuss the foundations related to this work, and analyze its challenges to motivate this study.

\begin{figure*}[!htb]
	\centering
	\includegraphics[width=0.60\textwidth]{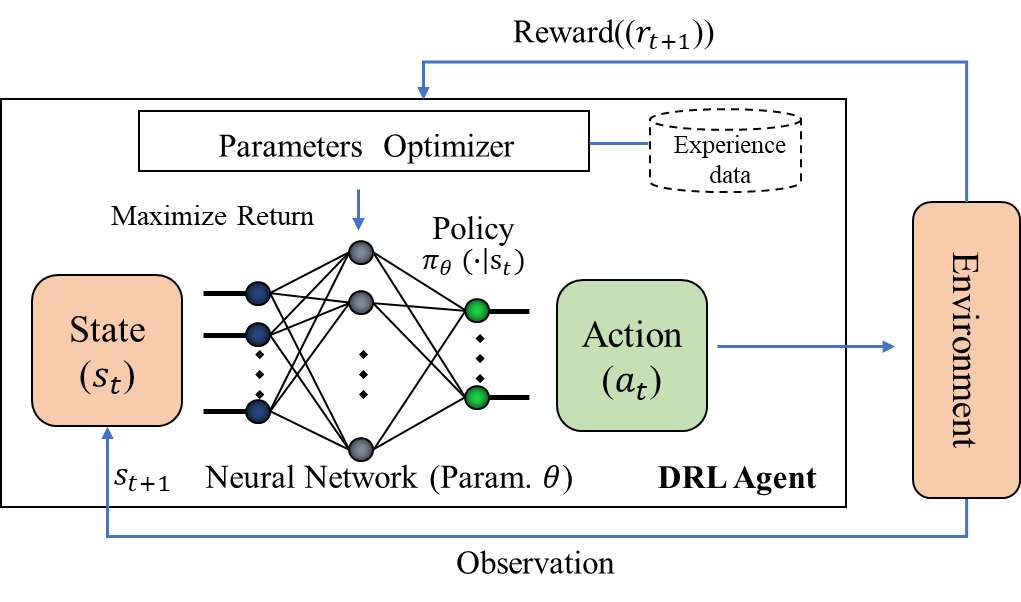}
	\caption{The semantics of deep reinforcement learning.}
	\label{fig:train}
\end{figure*}

\subsection{Preliminaries of deep reinforcement learning}
Formally, DRL can be modeled as a Markov Decision Process (MDP). More specifically, at each time step $t$, the DRL agent receives a state $s_t \in \mathcal S$ and makes its action decision $a_t \in \mathcal A$ according to a distribution called policy $\pi(\theta |s_t)$, where $\mathcal S$ and $\mathcal A$ are the state space and action space, respectively. The policy is basically modeled by a neural network with parameters $\theta$. After the action execution, the agent receives a reward $r_{t+1}$ and transitions to the next state $s_{t+1}$ according to the reward function $R(s_t,a_t,s_{t+1})$ and transition distribution $P(s_{t+1}|s_t,a_t)$, respectively. The return starting from time $t$ to the end of the interaction is defined by $G_t = \sum_{i=0}^{\infty}\gamma^i r_{t+i} $, where $\gamma\in[0,1]$ is the discount factor. The semantics of DRL are illustrated in Fig. ~\ref{fig:train}. DRL agents interact with the environment to collect experience data, which is usually denoted by $(s_t,a_t,s_{t+1},r_{t+1})$. Based on experience data, the agent is trained to learn an optimal policy for maximizing the return by updating the parameters $\theta$ of the neural network.

DRL can be classified to \textit{model-based} and \textit{model-free}. In \textit{model-based} variants, the transition and reward models (also named MDP dynamics) are known or learned first. Then the policy optimization is based on the models.  %\textcolor{red}{Due to the scope of this paper, we focus on \emph{model-free} deep reinforcement learning.}
There are mainly two categories in \textit{model-based} DRL~\cite{moerland2023model}. First is \textit{model-based DRL with a learned model}. This category of methods learns both the model and the global policy.  Examples are Dyna-style algorithms such as MB-MPO\cite{clavera2018model} and ME-TRPO\cite{kurutach2018model}. These algorithms learn the model through data from interaction with environments. Then model-free algorithms are utilized to learn the global policy based on the data generated by the learned model. Second is \textit{model-based DRL with a known model}. This category of methods plans over a known model, and only performs learning for the global policy. Examples are dynamic programming\cite{sutton2018reinforcement} and  Monte-Carlo tree search based algorithms (AlphaZero\cite{Silver2017}).

In contrast, \textit{model-free} DRL follows trail-and-error patterns and learns the policy directly. Basically, there are mainly three categories in \textit{model-free} DRL. First is \textit{value-based methods}. This category learns a value function that represents the expected return for being in a state or taking an action in the state, e.g., $V_\pi(s)$ and $Q_\pi(s,a)$. Then an optimal policy can be obtained based on the value function. Value-based methods are suited for discrete action spaces and deterministic policies. Second is \textit{Policy-based methods}. This category learns the policy directly and extracts actions according to the policy. Since the policy is actually a distribution of possible actions, continuous action spaces, and stochastic policies can also be learned. Compared to value-based methods, policy-based methods have better convergence properties. However, high variance and sample inefficient are common issues in policy-based methods. Third is \textit{Actor-critic based methods}, which integrates the merits of value-based and policy-based methods. Two separate neural networks are used in the architecture, named actor and critic. The actor learns the optimal policy and chooses the action by using the policy-based methods. The critic evaluates the selected action through the value-based methods. Due to the superiority in performance, actor-critic based methods have become more and more popular in recent years, e.g., PPO\cite{schulman2017proximal}, SAC\cite{haarnoja2018soft} and TD3\cite{fujimoto2018addressing}.

Alongside supervised learning and un-supervised learning, reinforcement learning performs optimization based on the experiences collected from interacting with uncertain environments. Finding a balance between \textit{exploration} (to discover new features about the world) and \textit{exploitation} (to utilize what it knows) is fundamental to the efficiency of reinforcement learning methods.
Meanwhile, according to the uniformity of the behavior policy and the target policy, DRL algorithms can be divided into \emph{on-policy} and \emph{off-policy}. Here, the behavior policy represents the policy used to generate the training data by interaction, while the target policy refers to the policy that needs to be learned. More specifically, in \emph{off-policy} learning, the behavior policy and the target policy are different. The goal of the off-policy learning is to learn the target policy by using the training data generated by the behavior policy. This has many benefits such as learning from demonstration, continuous exploration, and better sample utilization. However, it suffers from convergence and stability issues~\cite{fu2019diagnosing,kumar2019stabilizing}. In comparison, the behavior policy and the target policy are the same in \emph{on-policy} learning. As a result, better stability can be achieved in learning but sacrificing sample efficiency\cite{haarnoja2018soft}.

\subsection{Distributed training for deep learning}
In recent years, distributed training acceleration for deep learning has become a hot research topic and various methods have flourished in this field. Due to the tight connection between DL and DRL, knowledge and progress of distributed DL methods play a foundational role in parallel and distributed DRL. Basically, there are mainly three parallelism schemes of training acceleration for deep learning: data parallelism, model parallelism, and pipeline parallelism.   %Fig. X illustrates the diagrams of these parallelism schemes. 

Data parallelism~\cite{chu2006map,zinkevich2010parallelized,NIPS2014_1ff1de77,10.1145/3035918.3035933,zhang2020autosync} is the first proposed and the most common parallelism schemes. It partitions the training dataset (samples) into multiple subsets and dispatches them to multiple workers (e.g., compute nodes and devices). Each worker maintains a replica of the entire neural network model along with its local parameters. During training, each worker processes its subset of data independently and synchronizes parameters or gradients of the model across different workers either in a synchronous or asynchronous manner. This is the main parallelism scheme utilized in distributed DRL. Details can be found in Section~\ref{sec:sync}. 

Model parallelism~\cite{dean2012large,lee2014model,jain2020gems,narayanan2021efficient} slices the neural network model into disjoint partitions and assigns them to multiple workers for computation.  Compared to data parallelism, the computations between workers are no longer independent. Neurons with connections across workers require data transfers before continuing the computation to the next layer.  Due to the high computation dependency in neural networks, model parallelism is non-trivial for accelerating the training~\cite{Tang2020}. The main advantage of model parallelism is making training a large model, which cannot fit into the memory of one node, become possible.

Pipeline parallelism~\cite{Huang2019,Narayanan2019,narayanan2021memory,Luo2022} divides the neural network model into stages (consisting of consecutive layers) and assigns the stages to different workers. Each worker performs the computation of its stage for microbatches of samples one by one in a pipeline fashion.  This scheme can significantly improve parallel training throughput over the vanilla model parallelism by fully saturating the pipeline. Basically, pipeline parallelism is a special form of model parallelism. 

Combining these parallelism schemes therefore giving full play to their respective advantages is a promising direction\cite{Krizhevsky2014,Wu2016,NEURIPS2018_3a37abde,Jia2019,Rajbhandari2020,park2020hetpipe,Bian2021}. Krizhevsky et al.~\cite{Krizhevsky2014} propose a hybrid scheme of using data parallelism for convolutional layers and model parallelism for densely-connected layer. Wu et al.\cite{Wu2016} propose to make use of data and model parallelism to accelerate the training of recurrent neural networks. Rasley et al. propose ZeRo~\cite{Rajbhandari2020}, a parallelized optimizer based on model and data parallelism, which can train large models with over 100 billion parameters. Park et al. propose HeiPipe\cite{park2020hetpipe} that combines pipeline parallelism with data parallelism for training large deep networks.
%Jia et al. propose FlexFlow\cite{Jia2019}, a deep learning engine that automatically finds fast parallelism strategies beyond data and model parallelisms. 

\subsection{Challenges of parallel and distributed training for DRL}

Although above distributed training methods can serve as good references for training acceleration in DRL, copying these methods to DRL directly is infeasible. That is because there are differences in training schemes between DL and DRL. DL trains from supervised training datasets and seeks to minimize the error between the neural network model and the labeled data. In comparison, DRL does not have labeled data. It learns an optimal policy that leads to the maximum outcome through interacting with the environment. This will unavoidably cause differences in training acceleration between DL and DRL and trigger requirements of new techniques and methodologies. Overall, the challenges of parallel and distributed training for DRL can be summarized as follows.

\begin{itemize}
	\item Training DRL agents in parallel is a complex and dynamic problem that many components, such as actors, learners, and parameter servers, need to operate cooperatively. How to organize the architecture of these components and allocate the training tasks among them so as to maximize the system throughput is a challenging problem\cite{Petrenko2020,Espeholt2018}.
	\item Requiring to interact with environments (mostly simulated) to generate training samples is a distinguished feather of DRL. Training DRL agents for complex applications often need millions or even billions of experience samples. How to parallel simulations so as to increase the sample efficiency is demanding \cite{Makoviychuk2021,baker2020emergent}. 
	\item The workloads in distributed DRL training are heterogeneous, which may cause data movement across devices even nodes frequently. This will degrade the overall throughput and computation efficiency. How to saturate the computation resources by using different hardware architectures and corresponding computing parallelism techniques accordingly is challenging\cite{machupalli2022review}.
	\item Since the parallel learning machines may process at different speeds in a heterogeneous environment, obsolete gradients exist and can have a large impact on stability and convergence in training. How to aggregate the gradients and synchronize the model maintained in parallel workers is also non-trivial\cite{Chen2021a}.
	\item The essence of dominant training technology in distributed DRL is stochastic gradient descent based on backpropagation. However, this technology still suffers from local optima and expensive computational cost \cite{such2017deep}. How to exploit other optimization or search techniques in DRL to enable more sufficient training is desired.
\end{itemize}

In addressing these challenges, a profusion of studies have been proposed in the field of DRL acceleration using parallel and distributed. There is a necessity to collect, classify, and compare these studies in a structured manner, thereby facilitating researchers understanding the development of this field.
In the following sections, we survey the research for parallel and distributed DRL training and demystify the details of the key technologies. This includes system architectures for distributed DRL, simulation parallelism, computing parallelism, distributed synchronization mechanisms (\emph{backpropagation-based training}), and deep evolutionary reinforcement learning (\emph{evolution-based training}). In addition, we provide an overview and comparison of current open-source libraries and platforms that put parallel and distributed DRL into practice. 

\begin{figure*}[!htb]
	\centering
	\subfigure[Centralized architecture]{
		\label{fig:arch:sub1} %% label for first subfigure
		\includegraphics[width=0.38\textwidth]{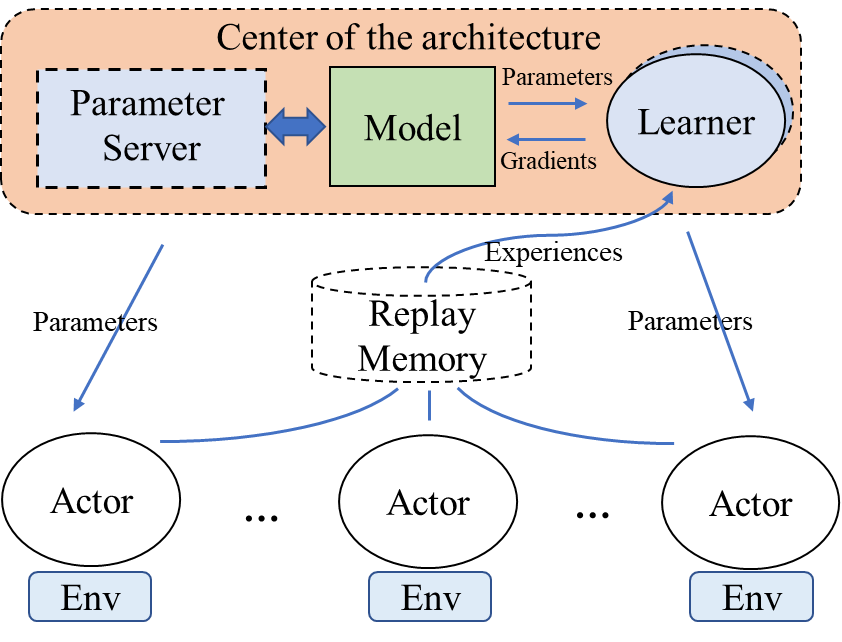}}
	\subfigure[Decentralized architecture]{
		\label{fig:arch:sub2} %% label for first subfigure
		\includegraphics[width=0.44\textwidth]{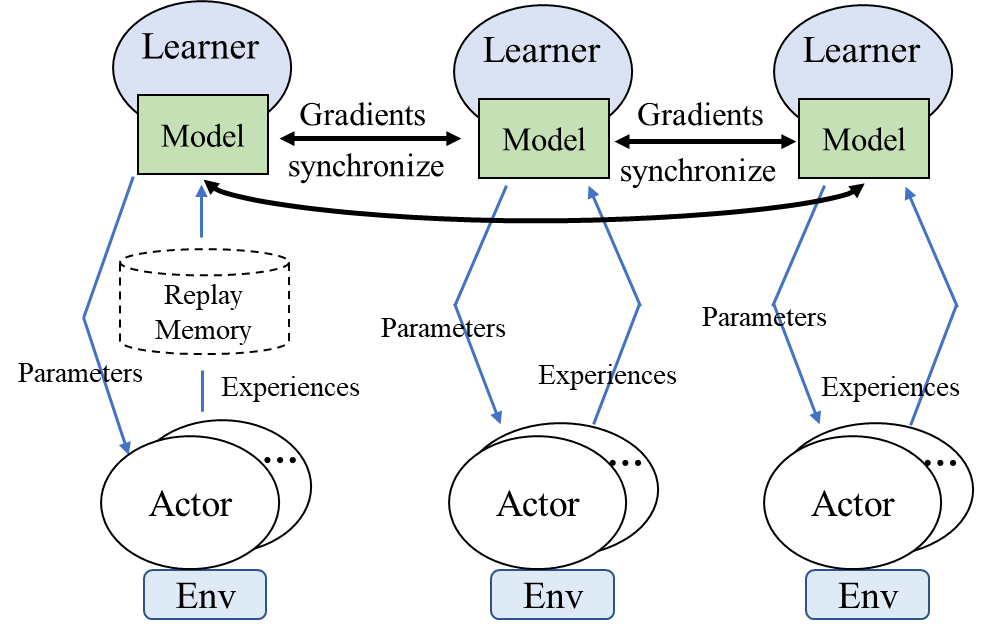}}
	\caption{Comparison of centralized and decentralized architectures for parallel and distributed training in DRL.}
	\label{fig:arch}
\end{figure*}

\section{System Architectures for Distributed DRL}\label{sec:arch}
Training DRL agents in parallel is a system engineering problem that many components need to operate cooperatively. %How to organize the architecture of these components and allocate the training tasks among them so as to maximize the system throughput is a challenging problem. 
Taking a panoramic view of the parallel and distributed DRL acceleration frameworks that bloomed in recent years, four components can be abstracted in the general case: 
\begin{itemize}
	
	\item \textit{Actor}: is in charge of interacting with the environment, including performing the actions obtained according to the policy, getting information such as observations and rewards, and producing experience data. %Since existing simulation environments basically can only be run on CPUs (e.g., OpenAI Gym and Atari game), Actors are usually distributed on a cluster of nodes withs CPUs to parallelize the computing tasks. 
	
	\item \textit{Learner}: collects or samples the experience data, and computes the gradients for updating the model. Since the gradient computation is usually based on a batch of experiences, high speed hardware like GPU is commonly used in Learners. 
	
	\item \textit{Parameter Server}: Maintains the up-to-date parameters of neural network models for Actors or Learners. Note that in some approaches (e.g., APE-X\cite{Horgan2018} and R2D2\cite{kapturowski2018recurrent}), the job of the parameter server is concurrently taken by the Learner.  %Since the network topology of the policy model for the agent in DRL is basically not changed during training, the model is determined with the parameters.
	
	\item \textit{Replay memory}: Stores the experience data produced by Actors. This exists in cases of off-policy reinforcement learning, e.g., Gorila\cite{Nair2015}, Ape-X\cite{Horgan2018}, R2D2\cite{kapturowski2018recurrent}, etc. 
	
\end{itemize}

According to the organization of these components, we classify the parallel and distributed training architectures for DRL into two classes: centralized and decentralized architectures.

\begin{figure*}[!htb]
	\centering
	\subfigure[Gorila\cite{Nair2015}]{
		\label{fig:arch:a} %% label for first subfigure
		\includegraphics[width=0.32\textwidth]{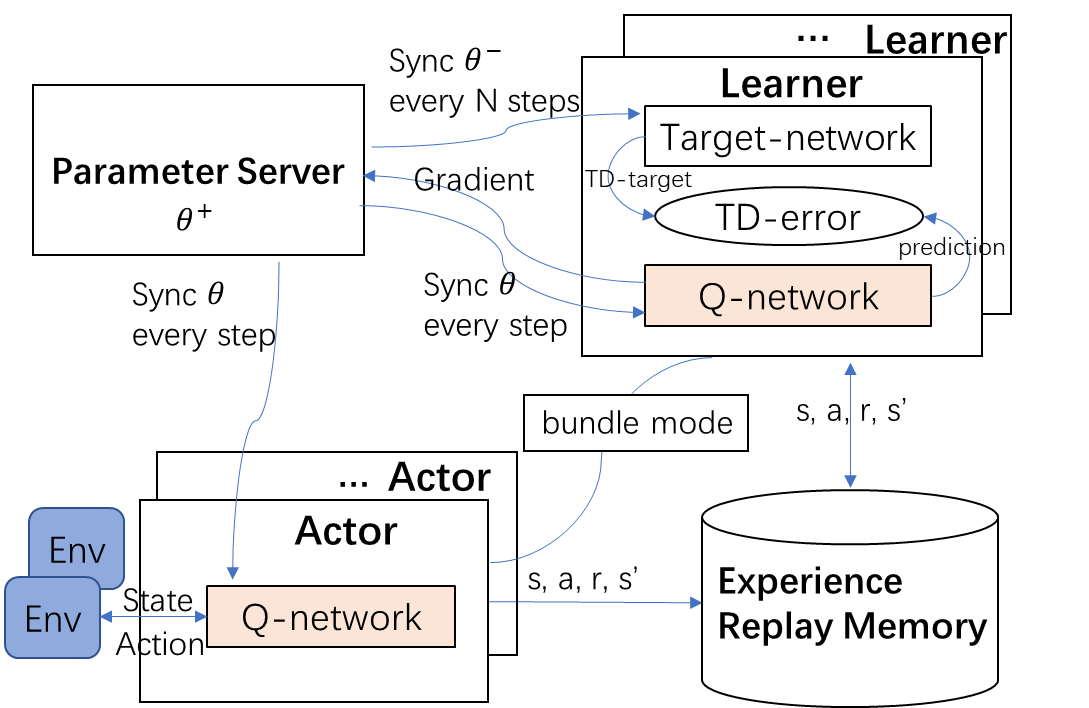}}
	\subfigure[APE-X\cite{Horgan2018}]{
		\label{fig:arch:b} %% label for first subfigure
		\includegraphics[width=0.32\textwidth]{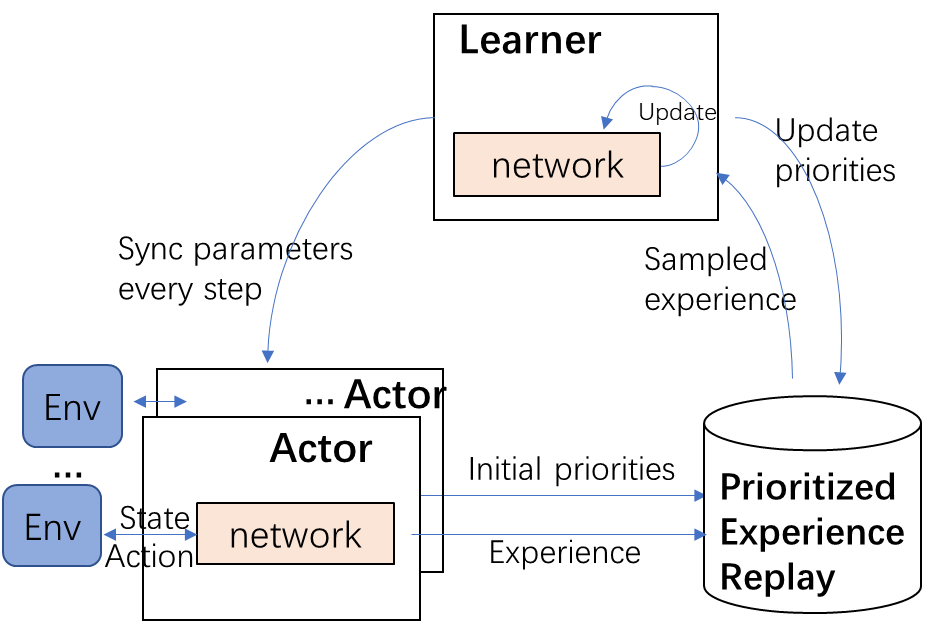}}
	\subfigure[A3C\cite{Mnih2016}]{
		\label{fig:arch:c} %% label for first subfigure
		\includegraphics[width=0.32\textwidth]{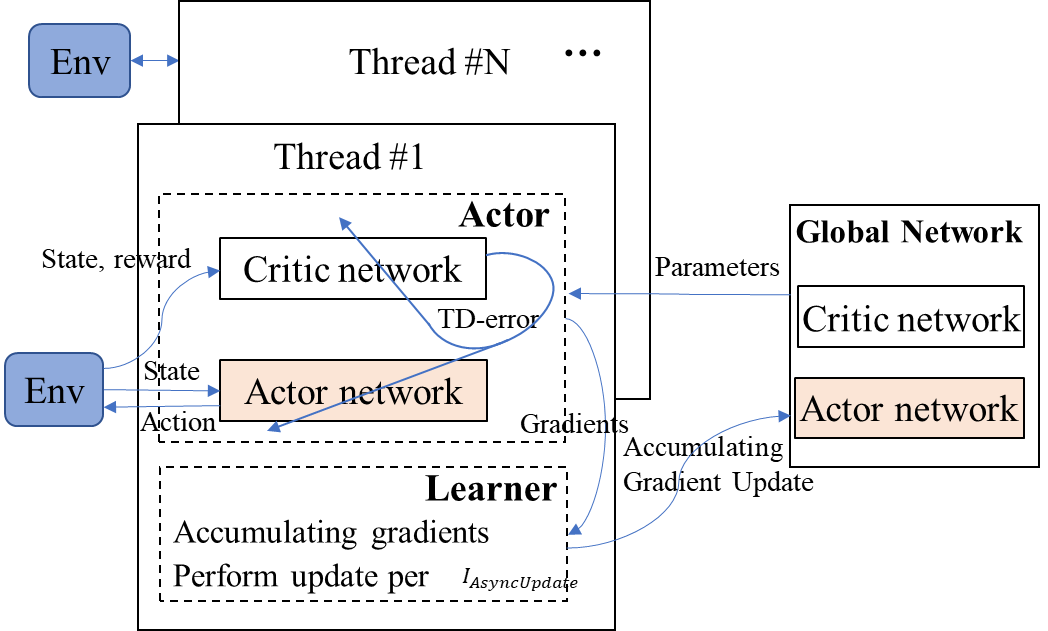}}\\
	\subfigure[IMPALA\cite{Espeholt2018}]{
		\label{fig:arch:d} %% label for first subfigure
		\includegraphics[width=0.32\textwidth]{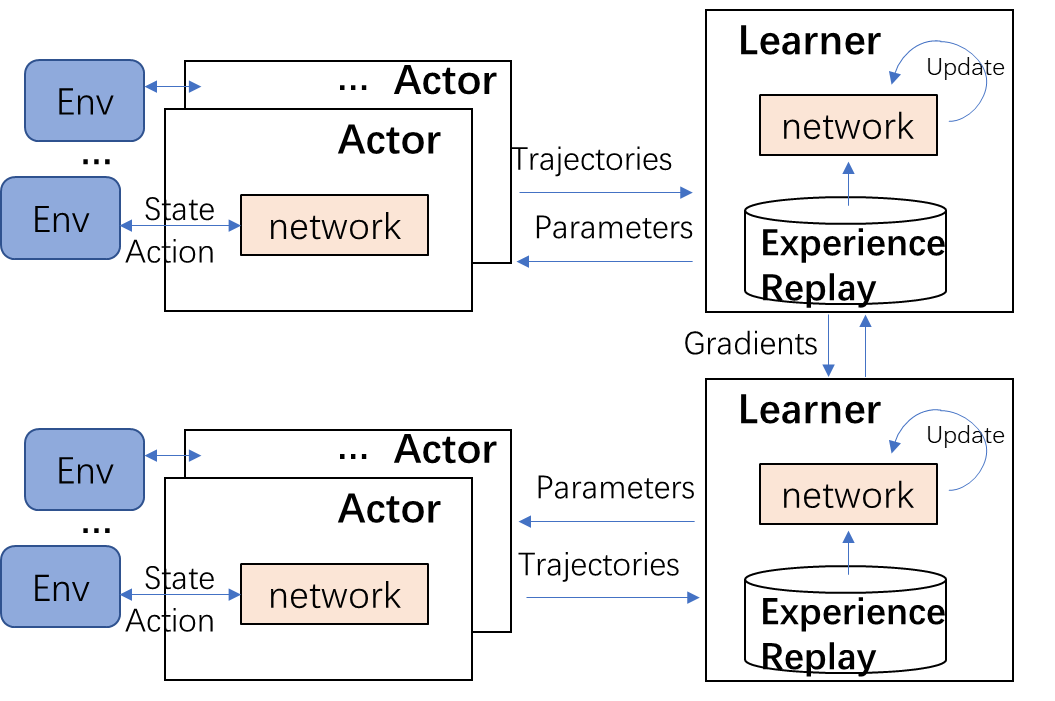}}
	\subfigure[rlpyt\cite{Stooke}]{
		\label{fig:arch:e} %% label for first subfigure
		\includegraphics[width=0.32\textwidth]{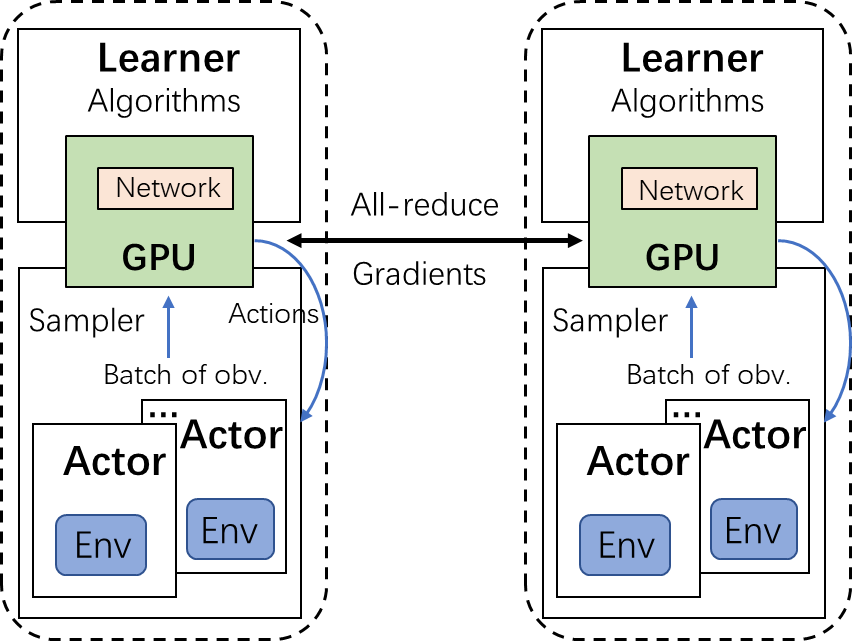}}
	\subfigure[DD-PPO\cite{Wijmans2019}]{
		\label{fig:arch:f} %% label for first subfigure
		\includegraphics[width=0.32\textwidth]{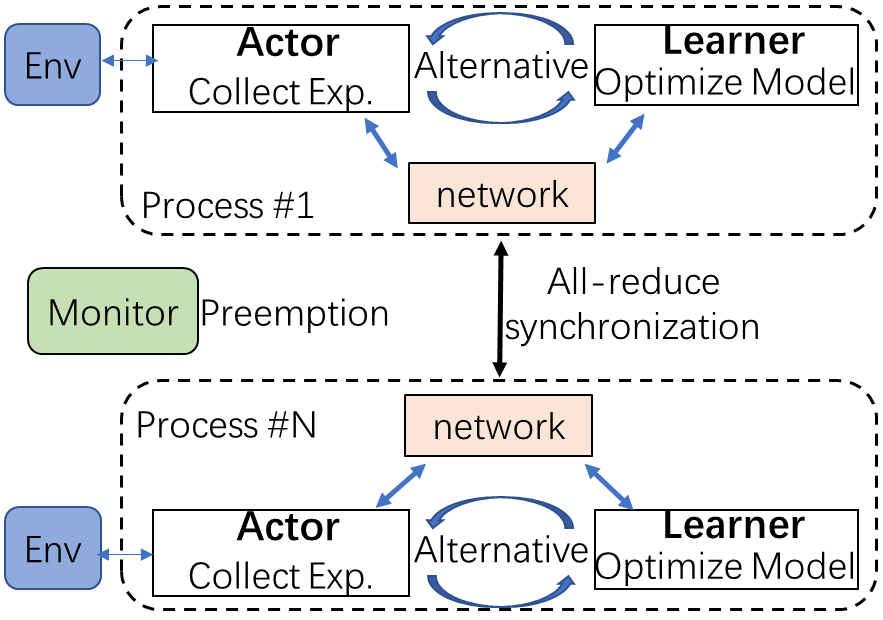}}
	\caption{Architectures of current parallel and distributed DRL methods.}
	\label{fig:archcurrent}
\end{figure*}
\subsection{Centralized architecture }
%In the centralized architecture, the Parameter Server is the center of the communication, as shown in Fig. XX(a). Each worker (including Actor and Learner) communicates to the Parameter Server to obtain the up-to-date parameters or transfer gradients in each iteration. More specially, Actors obtain the parameters from the Parameter Server to update its local model and send the experiences generated by interacting with environment to the corresponding Learner (or the Replay Memory if exist). Learners collect the experiences from its corresponding Actors (or samples from the Replay Memory with priorities~\cite{Horgan2018,kapturowski2018recurrent}), send its gradients computed latest to the Parameter Sever, and also synchronize the neural network model from the Parameter Server in some period or steps. Compared to Actors who usually have a large scale, the number of Learners can be one or many. One Learner scheme is common through leveraging batch processing unit like GPU. In any events, since a large number of workers as well as iterations could be involved in distributed DRL training, the bottleneck of the center node will have a strong impact on the training scalability.

In the centralized architecture, there is a center that maintains the global model of the neural network for learning. Learners optimize the global model by computing gradients based on the experiences from Actors. Both Learners and Actors synchronize their local model from the global model in some period or steps. Compared to Actors who usually have a large scale, the number of Learners can be one or many.  We abstract the general form of the centralized architecture as shown in Fig. \ref{fig:arch:sub1}. The global model is either maintained by the Parameter Server or a Learner. More specifically, if there are multiple Learners in the architecture, the Parameter Sever serves as the center for updating the global model and distributing it to Actors and Learners. Therefore, the star communication topology is constructed in the centralized architecture. 
In any event, since a large number of workers as well as iterations could be involved in distributed DRL training, the bottleneck of the center node will have a strong impact on the training scalability.

Silver et al.~\cite{Nair2015} propose a massively distributed method for the DQN named Gorila (General Reinforcement Learning Architecture). To the best of our knowledge, this is one of the pioneering works in the area of distributed training for DRL. Gorila is based on the centralized architecture and consists of four components: \textit{Actors}, \textit{Learners}, the \textit{Parameter Server}, and the \textit{Replay Memory}, as shown in Fig. \ref{fig:arch:a}. The center of Gorila is the Parameter Server, which is in charge of updating and distributing the network parameters. There can be many Actors and Learners in Gorila. Each Actor has a replica of the Q network and is responsible for generating experiences to the Replay Memory through interacting with the environment. Each learner also has a replica of the Q network and is in charge of producing the gradients to the parameter server based on the experiences from the Replay Memory. With the help of parallel Actors and Learners, Gorila can significantly improve the training performance for DQN, reaching a reduction of ten times in training duration on most games of Atari. 

Some variants over Gorila have also been proposed to further improve the performance. Horgan et al.~\cite{Horgan2018} propose a distributed method in centralized architecture for DQN with prioritized experience replay named APE-X. Rather than sampling uniformly in Gorila, APE-X focuses on learning the prioritized experiences with larger absolute temporal difference (TD) errors, as shown in Fig. \ref{fig:arch:b}. Further, a distributed method named Recurrent Replay Distributed DQN (R2D2)\cite{kapturowski2018recurrent} is proposed based on APE-X, which imports the recurrent neural network (LSTM) in training to achieve better performance in the partially observed environment. These two methods are based on a similar architecture to Gorila. However, only one Learner is used in these two methods, which is running on a GPU. By doing this, the Learner can exploit batch computing advantages of GPU for gradient computation compared to Gorila.  Besides, they leverage the single Learner to maintain the latest parameters instead of the Parameter Server.

Unlike Gorila which is for off-policy learning, Minh et al.~\cite{Mnih2016} propose a new type of distributed training method named A3C for on-policy learning. The architecture of A3C is shown in Fig. \ref{fig:arch:c}. We can see that A3C is based on a centralized architecture. Different from the aforementioned methods, A3C does not utilize the Replay Memory. Instead, A3C leverages the parallelization of Actors to enrich the diversity of environmental interaction experiences and accumulates updates over multiple steps to improve the stability of learning. Besides, the Actor and Learner are encapsulated together in an actor-learner thread. Further, although there is no Parameter Server, a global network is also maintained on a worker, which is considered as the center of the architecture. Each actor-learner thread sends accumulated gradients to and obtains the latest network parameters from the global network. 

\subsection{Decentralized architecture }
In the decentralized architecture, there is no central node that maintains the global model. Basically, more than one Leaner are deployed. Each Learner not only computes the gradients based on the experiences from the corresponding Actors, but also obtains the gradients from other learners. Then, each Learner updates its model by aggregating all the gradients. Here, All-reduce~\cite{liang2018rllib,Li2019} distributed communication mechanism is utilized to aggregate the gradients, which generally requires a fully connected communication topology for parallel workers.  Other than interacting with environments and producing the experiences, Actors update the parameters from Learners. We abstract the general form of the decentralized architecture as shown in Fig. \ref{fig:arch:sub2}. Compared to the centralized architecture, by using multiple Learners to aggregate the gradients and maintain the model, the decentralized architecture removes the communication congestion on the Parameter Server~\cite{Li2019}. 

A decentralized architecture of multiple learners with importance weighted is proposed in IMPALA~\cite{Espeholt2018}. In IMPALA, actors utilize CPU cores to interact with the environment, while learners are deployed on GPUs to give play to the gradient computation. To maintain the latest model in the whole system, gradients are aggregated across multiple learners synchronously and actors obtain the latest parameters from learners. 
Stooke et al.~\cite{Stooke2018,Stooke} propose an acceleration method of utilizing multiple GPUs for DRL named \textit{rlpyt}, which is based on the decentralized architecture. There are multiple Learners in the architecture. Each of them is running on a GPU and is in charge of data sampling and model inference. Besides, All-reduce is performed in each iteration to aggregate the gradients so as to keep the model in every Learner identical. This method has many variants that are applied with different DRL algorithms including A3C, PPO, DQN, etc. A similar architecture is adopted in DD-PPO\cite{Wijmans2019}, which is also decentralized and leverages multiple GPUs. Other than the method in~\cite{Stooke2018} that synchronizes the models of different GPUs in one machine, DD-PPO extends the architecture across many machines and has better scalability. Similar to A3C\cite{Mnih2016}, DD-PPO encapsulates the work of the Actor and Learner together in a worker. In this way, each worker alternates the workloads of experience collection, gradient aggregation, and optimization to achieve better resource utilization.

\subsection{Discussion}
These two classes of architectures are commonly used in recent studies. In the centralized architecture, it is known that the central node may encounter communication congestion that limits the scalability of distributed training \cite{verbraeken2020survey}. However, its merits on synchronization efficiency and realization simplicity are non-negligible. This is because the global updated model is maintained in a central node, from which all workers can easily obtain and keep consistent. In the decentralized architecture, the scalability issue is relieved by eliminating the single-point issue. Nevertheless, the models optimize separately and need to be synchronized through assembling, which may incur convergence latency and communication overhead ~\cite{Li2019}.

Overall, there are many open issues and emerging topics from the perspective of the learning system architecture. Firstly, the question of whether Actors should conduct model inference to derive actions remains unresolved. In scenarios where Actors are tasked with model inference,  Actors need to maintain the up-to-date copy of the model, obtain the actions through model inference, and transfer the experiences after environment interaction (e.g., IMPALA shown in Fig.~\ref{fig:arch:d}). Conversely, in other approaches, Actors focus on the interaction with the environment and give up the work of model inference (e.g., rlpyt shown in Fig.~\ref{fig:arch:e}). Instead, Learners obtain the actions through model inference and transfer them to Actors. %After that,  Actors return the interaction experiences to  Learners for policy learning.
Both paths of solutions have two sides. The former path takes full advantage of computational capacities in Actors to share the responsibilities of model inference. However, frequent model update requests from many Actors will significantly decrease the learning efficiency of Learner~\cite{yazdanbakhsh2020menger}. The latter path reduces the overhead of model synchronization among Actors and Learners, but it may incur latency on transferring actions and experiences \cite{espeholt2020seed}.

Besides, unlike the typical design of the system architecture where Actor and Leaner are the two types of distributed workers, some methods innovatively extend the workers to fine-grained types. For example, there are \textit{Actor worker}, \textit{Policy worker}, and \textit{Trainer} in \cite{mei2023srl}. The \textit{Actor worker} is responsible for environment interaction, the \textit{Policy worker} takes charge of policy inference, and the \textit{Trainer} is in charge of model learning. The main purpose of this design is to increase computational efficiency by allocating different workers to different computing nodes that suit their tasks, e.g., CPU, GPU or TPU nodes. A similar design can be found in SampleFactory \cite{Petrenko2020}, where there are \textit{Rollout worker}, \textit{Policy worker}, and \textit{Leaner} for the fine-grained types of parallel workers. Also, Li et al.~\cite{li2023parallel} refines the workers to \textit{Actor}, \textit{V-learner}, and \textit{P-learner} for data collection, value learning, and policy learning, respectively. Since there is a model learning job for Dyna-style \textit{model-based} DRL, Zhang et al.~\cite{zhang2020asynchronous} refine the parallel workers to \textit{Data Collection Worker}, \textit{Model Learning Worker}, and \textit{Policy Improvement Worker} in their asynchronous \textit{model-based} training method.

Furthermore, as previously discussed, the centralized architecture adheres to a star communication topology, whereas the decentralized architecture employs a fully connected communication topology. Each of these solutions carries its own drawbacks, such as the single point of failure in the centralized architecture and the increased synchronization overhead in the decentralized one. To achieve a better trade-off, Assran et al. \cite{assran2019gossip} propose gossip-based Actor-Learner architecture where parallel workers are organized in a peer-to-peer communication topology. Gossip~\cite{Mathkar2017,assran2019gossip} communication mechanism is used to exchange the model update in this method. Each parallel worker only communicates with its neighbors, and does not need to communicate with everyone else. The theoretical proof is given that the parallel workers' models are guaranteed to remain within $\epsilon-close $ during training. Experiments also show that this architecture achieves competitive performance compared to A3C\cite{Mnih2016} and IMPALA\cite{Espeholt2018}. Similarly, Sha et al. \cite{sha2022fully} propose distributed asynchronous policy evaluation based on directed peer-to-peer networks, allowing each parallel worker to update its value function locally by using data transferred from its neighbors.

\section{Simulation Parallelism}\label{sec:sim}
%simulation is importance. sample generation speed.
Requiring to interact with the environment to collect training samples is one of the main differences between DRL and DL.  
%Due to the differences in DRL applications, varieties of environments may face to in DRL training.
Simulations that simulate realistic environments are widely used to reduce training cost and time, especially for physics-based applications, e.g., robotics and autonomous driving. 
There are many simulation platforms that are popular in DRL training, e.g., OpenAI Gym, MoJoCo, Surreal, Unity ML, Gazebo, and AirSim.
%Basically, the simulation performance not only affects the feasibility of transferring the learned policies to physical world, but also plays an important role on the sample efficiency. 
Basically, the computations in simulations (e.g., physics calculation, rewards calculation, and  3D rendering) are involved in each step of training iteration. As the size of experiences increased for training complex applications, the simulation efficiency becomes more and more important in the training acceleration. For instance, 2.5 billion frames of experience are needed for PointGoal navigation in 3D scanned environments\cite{Wijmans2019}. How to parallel simulations so as to increase the sample efficiency is challenging. To this end, many approaches have been proposed. According to existing reinforcement learning literature, there are mainly two simulation parallelism strategies. 

\begin{figure*}[!htb]
	\centering
	\subfigure[CPU simulation parallelism pipeline]{
		\label{fig:sim:sub1} %% label for first subfigure
		\includegraphics[width=0.55\textwidth]{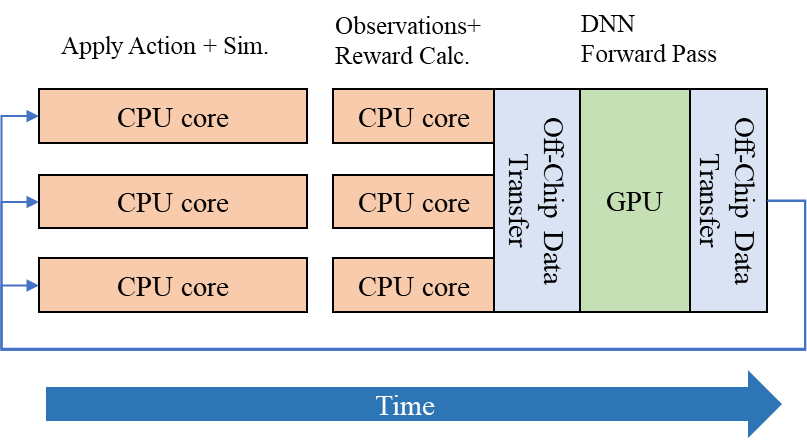}}
	\hspace{5mm}
	\subfigure[GPU simulation parallelism pipeline]{
		\label{fig:sim:sub2} %% label for first subfigure
		\includegraphics[width=0.28\textwidth]{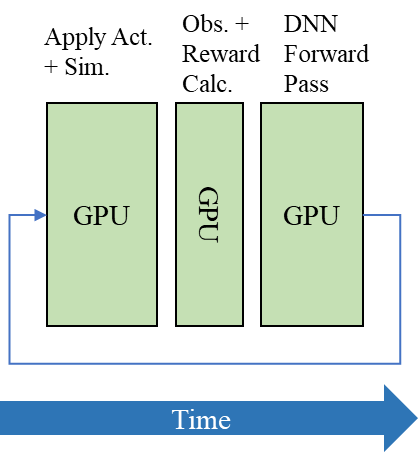}}
	\caption{Comparison of two simulation parallelism pipelines. (\emph{a}) The intermediate data needs to be copied from CPUs to GPUs back and forth during training in CPU simulation parallelism pipeline. (\emph{b}) GPU simulation of zero-copy enables directly accessing to simulation results in the GPU  buffers and keeps all of the computations on the GPU.~\cite{Makoviychuk2021}}
	\label{fig:sim}
\end{figure*}

%\begin{figure}[t]
%	\centering
%	\includegraphics[width=0.42\textwidth]{figures/terrain.jpg}
%	\caption{All agents (and task related objects) are loaded in a sharing scene~\cite{Liang2018}.}
%	\label{fig:sim:scene}       % Give a unique label
%\end{figure}

\subsection{Distributed simulation based on CPU clusters}
Many simulation platforms in use today are run on CPUs for many existing DRL methods in training, e.g., Atari in~\cite{mnih2015human,Mnih2016,Hessel2018,Gruslys2018}, MoJoCo in~\cite{schulman2015trust,lillicrap2015continuous,Schulman2017,fujimoto2018addressing} and OpenAI Gym in~\cite{Haarnoja2018,liang2018rllib,Tian2017}. A straightforward strategy for training acceleration is employing a cluster of CPUs to execute a number of instances of the environment in a distributed manner, where each instance normally runs on a process or a thread. Based on this, the experience data used for training can be generated at a higher speed. Besides, the experience data can also be more diversified by importing stochastic or using different policies when interacting with different instances of the environment. This strategy is adopted by many existing parallel and distributed DRL training methods such as Gorila~\cite{Nair2015}, A3C~\cite{Mnih2016}, APE-X~\cite{Horgan2018} and IMPALA~\cite{Espeholt2018}, and yields impressive results.  For example, in APE-X, approximately 50K environmental frames per second (FPS) can be generated by using 360 actor machines; each actor runs an instance of the environment by a CPU core.

Nevertheless, the scaling of these distributed simulation methods is determined by the number of CPU cores in the system. As the number of CPU cores and nodes in the cluster increases,  potential overhead may be incurred in communication across nodes~\cite{freeman2021brax,mittal2015timely}, synchronization~\cite{Wijmans2019,liu2016dynamic} and resource allocation~\cite{Dalton2020,guo2019limits}. Furthermore, as GPUs are commonly used in neural network computations, a combination of CPUs and GPUs is popular in most of the DRL researches. In this case, additional context-switching overhead would be introduced, as shown in Fig. \ref{fig:sim:sub1}. That is because intermediate data needs to be copied from CPUs to GPUs back and forth during training in this case. It is noted that simulation is only one part of the training platform. After a simulation step in the environment, the next state and a reward need to be computed. When using GPUs for neural network computations, the experience data (e.g., the next state and rewards) needs to be transferred from system memory to GPU memory for neural network inference. Then, actions are produced and need to be copied again to system memory for CPUs to perform the simulation step. 

\subsection{Batch simulation based on GPUs/TPUs}

In view of the limitations of CPU simulations on a large-scale, researchers have been committed to developing batch simulation methods based on specialized hardware architectures such as GPUs or TPUs. Liang et al.~\cite{Liang2018} propose a GPU-accelerated RL simulator to parallel the simulations, which can achieve thousands of humanoid robots concurrently simulated and generate 60K environmental frames per second in a single machine with one GPU and CPU.  
With the batch simulation framework, the humanoid robots can be trained to run in less than 20 minutes while using 1000 $\times$ less CPU cores compared to previous works~\cite{salimans2017evolution}. While relying on GPU to perform most of the simulation computations, this method still requires CPUs to perform tasks of getting state and applying controls. The performance of this method is also bounded by CPU-GPU communication bandwidth.

Therefore, zero-copy batch simulation that eliminates CPU-GPU communication has become a hot research topic in the field. Dalton et al.~\cite{Dalton2020} propose a CUDA Learning Environment named CulE to support GPU emulation for Atari. CuLE enables the Atari simulation to run directly within GPU memory, thereby eliminating off-chip communications that previous methods endured~\cite{Liang2018}. Up to 155K FPS in emulation only is achieved when emulated by CuLE with 4096 environments in parallel by 1 GPU. Further, other than targeting at  2D Atari simulation, Shacklett et al.~\cite{shacklett2021large} propose a large batch simulation method for 3D environments (PointGoal navigation).  This method bundles a large batch of requests for simulations and processes one entire batch at once. The key idea is to amortize the costs of simulation computation, synchronization, and communication across a batch of requests. Similarly, Makoviychuk et al.~\cite{Makoviychuk2021} propose a high performance robotics simulation platform named Isaac Gym for a variety of robotics environments. Isaac Gym offers a Tensor API that grants direct access to simulation results stored in GPU buffers, ensuring that all computations remain within the GPU. In this way, the bottleneck of off-chip communication between CPUs and GPUs can be eliminated, as shown in Fig. \ref{fig:sim:sub2}. Isaac Gym enables thousands of environments simulated on a single GPU and achieves up to 300$\times$ improvement in the training time over previous work~\cite{peng2021amp}.

Besides, Freeman et al.~\cite{Freeman2021} from Google Research propose an open-source library named Brax for large-scale rigid body simulation by using TPUs. Brax enables simulation computation and RL optimizer performed together on the same TPU. The experiment results show that Brax can achieve hundreds of millions of steps per second for MujoCo-ant on a TPUv3 8$\times$8 accelerator. 

\subsection{Discussion}

Currently, using a combination of CPUs and GPUs in DRL training is the mainstream in the field.  CPUs are used to simulate the environments and perform environmental interaction, while GPUs are used to perform neural network inference as well as weights update. Hence, CPU-GPU communication is an unavoidable performance bottleneck. To this end, zero-copy simulation that contains all the computation in GPUs or TPUs is one of the significant research directions for highly efficient simulations. This provides an alternative way of simulation parallelism, with no requirement of accessing to CPU clusters which most researchers find hard to get. Besides, different from previous works that run each environment instance in the individual simulation, sharing scenes with other robots in a simulation is an effective way to take advantage of batch parallelism and maximize the throughput~\cite{Liang2018,Makoviychuk2021,shacklett2021large}. Since texture and geometry objects tend to be large in size, naively loading these objects for every environment is unaffordable for GPU memory. For instance, Liang et al.~\cite{Liang2018} loads all agents (and task-related objects) in the same simulation. Shacklett et al.~\cite{shacklett2021large} maintain $K \ll N$ unique scenes in GPU memory, where $N$ is the number of parallel environments in a batch.  

Transferring the learned policies from simulation to reality is an essential demand for many real-world applications. Simulation-to-reality (Sim-to-real) is a topic of much interest. One straightforward strategy is to build a realistic simulator that can perfectly replicate the reality. Hence, the learned policies can be directly applied in real-world environments. This is referred as the \textit{zero-shot transfer}. In recent years, the \textit{zero-shot transfer} has been successfully demonstrated in several domains. Andrychowicz et al.~\cite{Andrychowicz2020} transfer learned policies for dexterous in-hand manipulation to physical robots by performing training entirely in simulation. Rudin et al.~\cite{rudin22a} demonstrate sim-to-real transfer for quadrupedal robots by using massively parallel DRL. Loquercio et al.~\cite{loquercio2021learning} achieve the \textit{zero-shot transfer} for high-speed UAV navigation while some realistic scenarios that were never experienced during training in simulation. The techniques for enabling seamless transfer in these approaches can be summarized as follows. First, add realistic sensor noise to the observations. Second, randomize physical properties in simulations such as friction coefficients and dynamics of objects. Third, learn with disturbances such as pushing the robot randomly and adding noise to rewards during training.

\section{Computing Parallelism}\label{sec:compu}
Other than simulation, intensive computing workloads are unavoidable in DRL training e.g., network inference, backpropagation, and evolutionary computation (see Section~\ref{sec:ec}). 
As a result, computing parallelism techniques are widely utilized to speed-up computations in DRL training. There are mainly three ways in the literature: cluster computing, single machine parallelism, and specialized hardware architectures (i.e., GPUs, FPGA, TPUs) acceleration. 

\subsection{Cluster computing}

Cluster computing usually includes multiple machines working together to perform computation-intensive or data-intensive tasks. Those machines are also called ‘nodes’ and inter-connected by a high-speed local network. Since the computing resources in individual nodes are relatively constrained, integrating the resources of multiple nodes is a straightforward and effective way for acceleration of large-scale processing.  In the early days when deep neural networks were not widely used in reinforcement learning, the classic cluster computing framework (i.e., MapReduce~\cite{dean2008mapreduce}) has been applied to reinforcement learning for acceleration~\cite{li2011mapreduce}. This method distributes the computation in the large matrix multiplication for Markov decision process (MDP) solutions such as policy evaluation and iteration. Considering the inadequacy of MapReduce for iterative computation involved in neural network training, Dean et al. propose DistBelief\cite{dean2012large} to utilize clusters of machines in training and inference for large-scale deep networks. However, this method targets deep network training, which is different from that of DRL. To this end, one of the earliest works on parallel and distributed DRL, Gorila~\cite{Nair2015}, is proposed. Gorila utilizes a cluster of machines to achieve an order of magnitude of reduction in training time than single-machine training. To the best of our knowledge, this is the enlightenment work that many methods, such as Ape-X\cite{Horgan2018}, IMPALA\cite{Espeholt2018}, and R2D2\cite{kapturowski2018recurrent}, spring up in this direction. From the evolution of related works, we can see that cluster computing is roughly a standard mode of accelerating training for DRL.   

% Table generated by Excel2LaTeX from sheet 'table'
\begin{table}[htbp]
	\footnotesize
	\centering
	\caption{Comparison of computing parallelism types in parallel and distributed DRL implementations. }
	\begin{tabular}{|p{8.3em}|p{2em}|p{2.4em}|p{2em}|p{2em}|p{2em}|p{14.6em}|p{16.5em}|}
		\toprule
		\multirow{2}{*}{Methods} & \multicolumn{5}{p{15em}|}{Computing parallelism Types} & \multirow{2}{*}{Implementation Details} & \multirow{2}{*}{Major Results} \\
		\cmidrule{2-6}    \multicolumn{1}{|c|}{} & CC   & MP/MT & \multicolumn{1}{p{2.4em}|}{GPU} & \multicolumn{1}{p{2.2em}|}{FPGA} & TPU   & \multicolumn{1}{c|}{} & \multicolumn{1}{c|}{} \\
		\midrule
		Gorila\cite{Nair2015}  & \checkmark     & \multicolumn{1}{r|}{} &       &       & \multicolumn{1}{r|}{} & 31 machines & 10× speedup over GPU implementation \\
		\midrule
		Ape-X\cite{Horgan2018}  &\checkmark    & \multicolumn{1}{r|}{} &  \checkmark &       & \multicolumn{1}{r|}{} & 360 CPU cores and 1 P100 GPU & 4× median scores over Gorila \\
		\midrule
		R2D2\cite{kapturowski2018recurrent}  &\checkmark    & \multicolumn{1}{r|}{} &  \checkmark &       & \multicolumn{1}{r|}{} & 256 actors and 1 GPU & 4× median scores over Ape-X  \\
		\midrule
		IMPALA\cite{Espeholt2018}  &\checkmark    &\checkmark    &  \checkmark &       & \multicolumn{1}{r|}{} & 500 CPU cores and 8 P100 GPUs & 250K FPS and multi-task setting \\
		\midrule
		Ray RLlib~\cite{liang2017ray}  &\checkmark    &\checkmark    &  \checkmark &       & \multicolumn{1}{r|}{} & 8,192 CPU cores on EC2 & completes training Mojoco in 3.7mins  \\
		\midrule
		ARS\cite{Mania2018}   &\checkmark    & \multicolumn{1}{r|}{} &       &       & \multicolumn{1}{r|}{} & 48 CPU cores on EC2 & 15× speedup over ES-based method\cite{salimans2017evolution} \\
		\midrule
		A3C\cite{Mnih2016}   & \multicolumn{1}{r|}{} &\checkmark    &       &       & \multicolumn{1}{r|}{} & 16 CPU cores  & 2× speedup over K40 GPU implementation \\
		\midrule
		Reactor\cite{Gruslys2018} &\checkmark    &\checkmark    &       &       & \multicolumn{1}{r|}{} & 20 CPU cores & 4× speedup over A3C \\
		\midrule
		DBA3C\cite{Adamski2018} &\checkmark    &\checkmark    &       &       & \multicolumn{1}{r|}{} & 64 nodes with 768 CPU cores  & completes training Atrai 2600 in 21 mins \\
		\midrule
		DPPO\cite{Heess2017}  & \multicolumn{1}{r|}{} &\checkmark    &       &       & \multicolumn{1}{r|}{} & 64 actors & \textgreater20× speedup over A3C  \\
		\midrule
		D4PG\cite{Radients2018}  & \multicolumn{1}{r|}{} &\checkmark    &       &       & \multicolumn{1}{r|}{} & 64 CPU cores & 4× higher return than PPO \\
		\midrule
		SampleFactory\cite{Petrenko2020} & \multicolumn{1}{r|}{} &\checkmark    &  \checkmark &       & \multicolumn{1}{r|}{} & 36 CPU cores and a 2080Ti GPU & 4× speedup over SEED\_RL \\
		\midrule
		GA3C\cite{Babaeizadeh2017}  & \multicolumn{1}{r|}{} &\checkmark    &  \checkmark &       & \multicolumn{1}{r|}{} & 16 CPU cores and 1 Titan X GPU & 45× speedup over A3C \\
		\midrule
		PAAC\cite{clemente2017efficient}  & \multicolumn{1}{r|}{} &\checkmark    &  \checkmark &       & \multicolumn{1}{r|}{} & 4 CPU cores and a GTX 980 Ti GPU & \textgreater6× speedup over Gorila \\
		\midrule
		rlpyt\cite{Stooke}\cite{Stooke2018} & \multicolumn{1}{r|}{} &\checkmark    &  \checkmark &       & \multicolumn{1}{r|}{} & 8 P100 GPUs and 40 CPU cores  & 6× speedup using 8 GPUs relative to 1 GPU \\
		\midrule
		Dactyl\cite{Andrychowicz2020} &\checkmark    &\checkmark    &  \checkmark &       & \multicolumn{1}{r|}{} & 384 nodes (6144 cores and 8 GPUs) & 5.5× speedup over implementation with 1 GPU and 768 CPU cores \\
		\midrule
		DD-PPO\cite{Wijmans2019} & \multicolumn{1}{r|}{} &\checkmark    &  \checkmark &       & \multicolumn{1}{r|}{} & 256 V100 GPUs  & 196× speed up over 1 V100 GPU \\
		\midrule
		MSRL\cite{zhu2023msrl} & \multicolumn{1}{r|}{} &\checkmark    &  \checkmark &       & \multicolumn{1}{r|}{} & 64 GPUs  & 3× speedup over Ray RLlib \\
		\midrule
		SRL\cite{mei2023srl} & \checkmark & \multicolumn{1}{r|}{}    &  \checkmark &       & \multicolumn{1}{r|}{} &  15K CPU cores and 32 A100 GPUs  & 5× speedup over OpenAI Rapid\cite{berner2019dota} \\
		\midrule
		SpeedyZero\cite{mei2023speedyzero} & \checkmark & \multicolumn{1}{r|}{}    &  \checkmark &       & \multicolumn{1}{r|}{} &  192 CPU cores and 20 A100 GPUs  &  mastering Atari benchmark within 35 minutes using only 300k samples. \\
		\midrule
		NNQL\cite{Su2017}  & \multicolumn{1}{r|}{} & \multicolumn{1}{r|}{} &       &  \checkmark & \multicolumn{1}{r|}{} & Arria 10 AX066 FPGA & 346× speedup over GTX 760 GPU \\
		\midrule
		TRPO\_FPGA\cite{Shao2017} & \multicolumn{1}{r|}{} & \multicolumn{1}{r|}{} &       &  \checkmark & \multicolumn{1}{r|}{} & Intel Stratix-V FPGA & 19.29× speedup over i7 CPU \\
		\midrule
		DDPG\_FPGA\cite{Guo2019} & \multicolumn{1}{r|}{} & \multicolumn{1}{r|}{} &       &  \checkmark & \multicolumn{1}{r|}{} & Intel Stratix-V FPGA & 4.53× speedup over i7-6700 CPU core \\
		\midrule
		FA3C\cite{Cho2019}  & \multicolumn{1}{r|}{} &\checkmark    &       &  \checkmark & \multicolumn{1}{r|}{} & Xilinx VCU1525 VU9P FPGA & 27.9\% better than Tesla P100 \\
		\midrule
		PPO\_FPGA\cite{Meng2020} & \multicolumn{1}{r|}{} &\checkmark    &       &  \checkmark & \multicolumn{1}{r|}{} & Xilinx Alveo U200 & 27.5× speedup against Titan Xp GPU \\
		\midrule
		On-chip replay\cite{Meng2022} & \multicolumn{1}{r|}{} &\checkmark    &       &  \checkmark & \multicolumn{1}{r|}{} & Xilinx Alveo U200 acceler & 4.3× higher IPS over GTX 3090 GPU \\
		\midrule
		AlphaZero\cite{Silver2017} &\checkmark    &\checkmark    &       &       &\checkmark    & 5000 TPUs v1 and 64 TPUs v2 cores  & defeats world-champion program by training within 24 hours \\
		\midrule
		AlphaStar\cite{vinyals2019grandmaster} &\checkmark    &\checkmark    &       &       &\checkmark    & 3,072 TPU v3 and 50,400 CPU cores & achieves above 99.8\% of ranked human players by training in 44 days \\
		\midrule
		OpenAI Five\cite{berner2019dota} &\checkmark    &\checkmark    &       &       &\checkmark    & 1,536 GPUs and 172,800 CPU cores & defeats Dota 2 world champion (Team OG) by training in 10 months \\
		\midrule
		GATO\cite{Reed2022}  &\checkmark    &\checkmark    &       &       &\checkmark    & 256 TPU v3 cores & handles 604 distinct tasks with a single network \\
		\midrule
		SEED\_RL\cite{espeholt2020seed} &\checkmark    &\checkmark    &       &       &\checkmark    & 520 CPU and 8 TPU v3 cores & 11× faster than the IMPALA with a P100 GPU \\
		\bottomrule
	\end{tabular}%
	\begin{tablenotes}
		\footnotesize
		\item CC: Cluster Computing; MP/MT: Multiprocessing or Multithreading; Statistics are collected from the corresponding papers. 
	\end{tablenotes}
	\label{tab:parallism}%
\end{table}% 	

\subsection{Single machine parallelism}

Other than cluster computing that uses multiple machines, single machine parallelism utilizes multiprocessors or multi-core CPUs in a single machine to increase the computing ability. The rationale for applying single machine parallelism is that distributed clusters may not be affordable for most researchers, unlike the widely available commodity workstations. Besides, data transfer, model synchronization across machines during training will also incur noticeable overhead. Multiprocessing and multithreading are the key technologies in single machine parallelism.  More specifically, multiprocessing refers to operating different processes simultaneously by more than one CPU processor, while multithreading refers to executing multiple independent threads in parallel by a single CPU, especially the multi-core CPU. 

The representative method is A3C~\cite{Mnih2016}. In A3C, many actor-learner threads are launched in parallel, and each thread performs a training procedure separately, including environment interaction, experience collection, and model update. In this way, the speed of data generation and the sample efficiency can be improved significantly. The results in~\cite{Mnih2016} showed that A3C using 16 CPU cores surpasses DQN variants using Nvidia K40 GPU in half of the training time and achieves comparative performance to Gorila which uses 100 machines. 

Petrenko et al. present a high-throughput training system named SampleFactory~\cite{Petrenko2020}, which is designed based on a single machine with a multi-core CPU and a GPU. Though optimizing the efficiency and resource utilization in a single machine setting, Sample Factory can achieve throughout as 130K FPS (Frame per Second) and 4 times speedup over the state-of-the-art baseline SEED\_RL~\cite{espeholt2020seed} with a workstation-level PC. One of the significant contributions of SampleFactory is that it allows large-scale DRL experiments accessible to a wider community by reducing the computational requirements. 

A combination of cluster computing and single machine parallelism is quite common in literature. Fiber~\cite{Zhi2020} extends multiprocessing from one machine to distributed environments. AlphaGo~\cite{Silver2016} uses asynchronous multi-threads to improve the performance in large-scale Monte Carlo tree search. IMPALA~\cite{Espeholt2018} provides single machine settings as well as distributed settings to accelerate large-scale DRL training. The benefit of single machine parallelism over cluster computing is removing the communication overhead of sending data across machines by keeping all the computing in a single machine. Note that the latency of synchronizing the model may have a noticeable impact on the stability of learning, especially for on-policy learning.

\subsection{Specialized hardware architectures}

Since there are lots of batch processing operations (e.g. matrix multiplication) in neural network training and inference, the specialized computation hardware architectures with large throughputs such as GPUs (Graphics Process Units), FPGAs (Field Programmable Gate Arrays), and TPUs (Tensor Process Units) are preferable. With massive ALUs (arithmetic and logic units) and good programmability, GPUs are utilized to accelerate in a wide range of applications that go well beyond traditional graphics processing. Comparably, FPGA supports customized operations for a specific application by user-programmable interconnects. Although FPGAs might not be superior to GPUs in computation efficiency, they are preponderant in reducing energy consumption. In addition, TPUs are custom-developed application-specific integrated circuits (ASICs) for machine learning workloads and achieve high throughput and low energy consumption for matrix computations. Overall, all these hardware architectures can exploit the inherent computing parallelism to achieve speedup on data-intensive computing tasks. 

In recent years, Nvidia has been devoting itself to utilizing GPUs in distributed DRL. Mohammad et al. propose GA3C~\cite{Babaeizadeh2017}, an architecture based on A3C~\cite{Mnih2016} with highlighted on GPU utilization to augment the training speed. GA3C employs trainer threads to collect batches of data and submit them to a GPU for exploiting the GPU’s computational capabilities better. Rather than utilizing a single GPU, a multi-GPU RL framework has also been proposed~\cite{Liang2018}, in which 32 GPUs in maximum are reported in the experiments. Based on that, hundreds to thousands of locomotion robots are parallelized to accelerate the training. Lasse et al. propose IMPALA~\cite{Espeholt2018} to solve large-scale, multi-task RL problems with a multi-GPU system (NVIDIA DGX-1), which contains 8 NVIDIA Tesla V100 GPUs.

FPGAs are utilized in tabular reinforcement learning for acceleration in the early days~\cite{da2018parallel,rajat2020qtaccel}. As the neural network has been induced in reinforcement learning, more room for improvement by FPGAs is created. Many works are proposed to use FPGAs in accelerating backpropagation in DRL. Su et al.~\cite{Su2017} propose an FPGA acceleration system for DQN, which allows dynamically reconstructing the network to achieve better learning results. Their experiment results show that the proposed system can achieve up to 346 times and 77 times speedup compared to GPU implementation and CPU implementation, respectively. Cho et al. propose a FPGA-based A3C Deep RL platform named FA3C~\cite{Cho2019}. In FA3C, simple and generic processing elements are designed for all types of DNN layers. Compute unit pairs are also proposed for inference and training tasks separately. Experiment results show that FA3C achieves 27.9\% higher IPS values (the number of inferences processed per second) and 1.62 times better energy efficiency than GPU implementation (Nvidia Tesla P100). 

TPUs are designed specifically for machine learning workloads and used frequently in the latest Google findings in the DRL domains such as AlphaZero~\cite{Silver2017}, AlphaStar~\cite{vinyals2019grandmaster}, GATO~\cite{Reed2022}, etc. In terms of distributed training platforms, Google research teams propose a scalable reinforcement learning framework called SEED\_RL~\cite{espeholt2020seed}, in which multi-TPUs architecture is utilized and the learning speed is significantly improved compared to the baseline based on multi-GPUs. For example, an 11 times speedup is achieved by SEED\_RL over a strong baseline IMPALA on DeepMind tasks.

Although these specialized hardware architectures have brought a huge speedup, CPUs also play an essential role in DRL training (e.g., environment interaction, task scheduling, and parameter sharing). A heterogeneous architecture of CPUs and specialized hardware is a promising endeavor in this area. Combining CPUs and GPUs has become the popular way of saturating the computation power of GPUs in recent years, and many works are proposed to utilize CPU-GPU heterogeneous architectures~\cite{Babaeizadeh2017,Espeholt2018,Stooke2018,mei2023speedyzero}. Other than this, Meng et al. propose a CPU-FPGA heterogeneous platform for accelerating Proximal Policy Optimization~\cite{Meng2020}. Rather than only focusing on specific RL algorithms, Meng et al., further propose a more generic CPU-FPGA accelerator  using on-chip replay management for widely used RL algorithms including DQN and DDPG~\cite{Meng2022}.  By allocating the workloads to CPU and FPGA sophisticatedly, CPU-FPGA heterogeneous method achieves significant improvement in overall throughput.

\subsection{Discussion}

Table \ref{tab:parallism} compares the computing parallelism types utilized in current parallel and distributed DRL implementations. %From the table we can see that As parallelism applications in DRL become more and more ambitious, the state-of-the-art solution only took 3.6 mins [ray] to train Atari 2600 games. In comparison, it took 15 days to train a single Atari game, which at that time was a significant breakthrough in human history [].  
We can see in the table that all of these parallelism types are widely used, and a combination of different computing parallelism techniques is popular. More specifically, all these three types of parallelism are reported in the latest research such as Ray RLlib\cite{liang2017ray}, SEED\_RL\cite{espeholt2020seed}, Dactyl\cite{Andrychowicz2020}, AlphaZero\cite{Silver2017}, GATO\cite{Reed2022}, etc. As parallelism in DRL becomes more and more ambitious, achievements in training efficiency in this area have been proven. The state-of-the-art solution only took 3.6 mins~\cite{liang2017ray} to train Atari 2600 games. In comparison, it took 15 days to train a single Atari game several years ago, which at that time was a significant breakthrough in human history~\cite{mnih2015human}.

Besides, with the parallelization of computing resources, computing task scheduling plays an important role in enhancing the efficiency of distributed DRL systems. The computing workloads involved in distributed DRL training are diverse and heterogeneous, including environment interaction, network interference, gradient backpropagation, and model synchronization, which are handled by workers such as Actors, Learners, or the Parameter Server. %These computing workloads are deployed on difference workers such as Actors, Learners and the Parameter Sever.  
Hence, it is essential to effectively schedule these computing tasks across processors even machines with varying resource configurations, thereby minimizing the overall training time. There are many scheduling strategies have been proposed in current distributed DRL systems. 
\begin{itemize}
	\item Load balancing. This strategy enables load-aware scheduling that assigns computing tasks to distributed hardware, considering factors such as resource contention and input locality. Ray\cite{Ray18} utilizes fine-grained and coarse-grained load balancing methods for stateless and stateful computations, respectively. rlpyt\cite{Stooke2018} forms two alternating groups of simulation processes, and schedules GPU to serve each group in turn to keep utilization high. Meng et al. \cite{Meng2020} propose inter-load balancing method for two compute units (CUs) involved in training value and policy networks. Li et al.~\cite{li2023parallel} propose to explicit control the frequencies for parallel workers to achieve load balance. 
	\item Resource dynamic adjusting. This strategy dynamically scales the resources for workers to accelerate DRL training with minimum costs. MINIONSRL \cite{yu2024cheaper} leverages a scheduler that dynamically adjusts the number of Actor workers to optimize the DRL training with minimal training time and costs. Similarly, Meng et al. design a scheduling mechanism that re-allocates CPU threads to processing a sub-batch of training for the Learner in heterogeneous computing environments.
	\item Computation and communication overlapping. 
	%Although the computing and communication tasks can be executed in parallel 
	In distributed DRL training, computing tasks can be blocked by communication tasks due to their execution dependencies. For example, gradient aggregation has to wait for the completion of gradient transfer for parallel workers. This strategy schedules the computing and communication tasks to reduce the waiting time, thereby improving the computation utilization. PEARL~\cite{meng2024pearl}, an open-source distributed DRL framework, designs a Learner module that enables scheduling to overlay the communication and computation. Mei et al~\cite{mei2023speedyzero} propose an optimized data transfer scheduling technique that overlaps computation in distributed DRL training. 
	\item Preemptive scheduling. This strategy proactively terminates lagging tasks, as each worker must wait for all parallel workers to complete during synchronous training. Wijmans et al. propose a straggler preemption mechanism in DDPPO~\cite{Wijmans2019}, where the slow-running worker is preempted once a certain percentage of the other parallel workers are completed collecting their experience data in one iteration. Similarly, PyTorchRL~\cite{Bou2020} employs a comparable preemption mechanism within its distributed DRL training framework. 
\end{itemize}

\section{Distributed Synchronization Mechanisms}\label{sec:sync}

The dominant solution\footnote{Another promising training solution based on evolutionary learning will be discussed in Section~\ref{sec:ec}} for training acceleration in distributed DRL is employing a number of workers to cooperatively train the model based on \textit{distributed stochastic gradient descent (DSGD)}~\cite{Tang2020,Ben-Nun2019}. This is based on the data parallelism introduced in Section~\ref{sec:bg}. 
During DSGD training, each parallel work holds a copy of the model, performs training based on a subset of environmental experiences, and then aggregates the update to the target model cooperatively.  It is noted that distributed synchronization among workers is vital to the training efficiency, especially in the heterogeneous environment where the parallel workers may process at different speeds.  
%is responsible of collecting the gradients from a number of workers and produces a updated model by model synchronization for the next iteration. 
Variants of synchronization mechanisms are systemically studied in deep learning area, i.e., Bulk Synchronous Parallel (BSP)\cite{Pan2017}, Asynchronous Parallel (ASP)~\cite{Lian2018} and Stale Synchronous Parallel (SSP)~\cite{Ho2013}. These mechanisms lay a good technical foundation for distributed DRL.

\begin{figure*}[!htb]
	\centering
	\subfigure[Asynchronous Off-policy Training]{
		\label{fig:consist:sub1} %% label for first subfigure
		\includegraphics[width=0.45\textwidth]{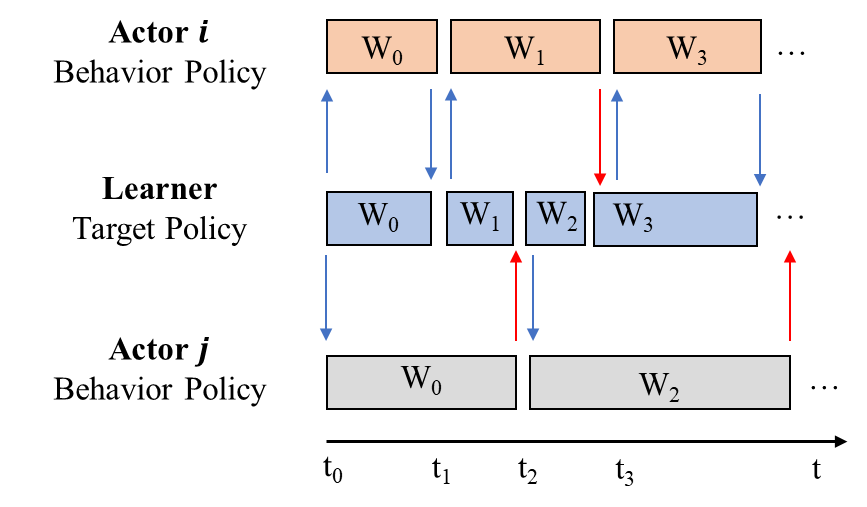}}
	\subfigure[Synchronous On-policy Training]{
		\label{fig:consist:sub2} %% label for first subfigure
		\includegraphics[width=0.45\textwidth]{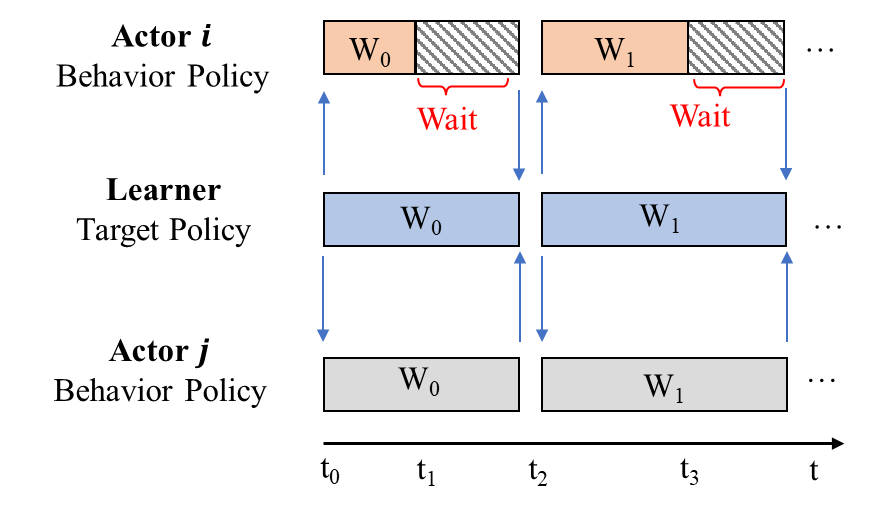}}
	\caption{Comparison between different synchronization mechanisms in distributed DRL. The rectangular bar represents the corresponding training step for the Actor and the Learner.  Arrows represent data transmission for the model parameter update or the rollout data.}
	\label{fig:consist}
\end{figure*}

However, since importing environment interaction and dividing the parallel workers into different roles such as Actors and Learners, DRL may face new difficulties while applying distributed synchronization approaches.  The model holding in Actors for the behavior policy may lag behind the model in Learners for the target policy after several iterations in parallel settings.  This makes the behavior policy different from the target policy gradually. It is noted that there are off-policy and on-policy schemes for DRL algorithms (see Section~\ref{sec:bg}). According to the training schemes and the way of synchronization among workers, there are mainly two types of distributed synchronization mechanisms in current distributed DRL solutions: asynchronous off-policy training and synchronous on-policy training.

\subsection{Asynchronous Off-policy Training}
Asynchronous off-policy training is the distributed synchronization mechanism that was initially applied in distributed DRL, and it has been widely used later on, e.g., Gorila\cite{Nair2015}, APE-X\cite{Horgan2018}, R2D2\cite{kapturowski2018recurrent} etc. 	
In the asynchronous training, the parallel workers operate independently. As shown in Fig.\ref{fig:consist}, each worker performs model update at its own pace without waiting for other workers. This will naturally lead to differences between the behavior policy model for Actors and the target policy model for Learners.  We can see in Fig.\ref{fig:consist:sub1}, at time $t_1$, the behavior policy with parameter $W_0$ for Actor $j$ is different from the target policy with parameter $W_1$ for the Learner. The same phenomenon can also be seen at time $t_2$ , when the behavior policy with parameter $W_1$ for Actor $i$ is different from the target policy with parameter $W_2$ for the Learner. Therefore, this type of synchronization mechanisms is mainly used for off-policy algorithms which do not require the consistency between the behavior policy and the target model while learning. DQN is a classic off-policy algorithm, and many asynchronous distributed DRL methods such as Gorila, APE-X, and R2D2 are based on DQN.
More specifically, asynchronous off-policy training is normally based on a centralized architecture and there is a Parameter Sever maintaining the global model. Each Actor pushes its interaction training data to the Replay Memory and pulls the up-to-date parameters from the Parameter Server asynchronously. The Learner updates the global model based on the training data in the Replay Memory. In this way, every worker tries to optimize the global model by contributing its efforts based on the local copy of the model, thereby accelerating the model training.

However, it is known that off-policy algorithms suffer from the stability issue \cite{fu2019diagnosing}. That is mainly because the distribution under the target policy shifts from that under the training data collected by the behavior policy, named \textit{policy-lag}~\cite{Espeholt2018}. To relieve this problem, there are many off-policy correction methods have been proposed in distributed DRL. Espeholt et al. propose \textit{V-trace} correction methods in IMPALA \cite{Espeholt2018}. In this method, \textit{V-trace targets} are computed based on trajectories collected by the behavior policy and the current target value function under the target policy, where two truncated importance sampling (IS) weights are imported to improve convergence. Babaeizadeh et al. propose $\epsilon$-correction in GA3C \cite{Babaeizadeh2017}, which adds a small constant $\epsilon$ during gradient estimation. This prevents the log value of the sampled action probability from becoming very small and leading to numerical instabilities.

Further, the nature of the asynchronous update will also cause \textit{stale update}, which occurs when a worker updates the target model using training data generated by the obsolete model~\cite{Pan2017}. As shown in Fig. \ref{fig:consist:sub1}, the Learner, Actor $i$ and Actor $j$ at time $t_0$ hold the model denoted as $W_0$. Actor $i$ at time $t_1$ pushes the training data to create a new model denoted as $W_1$. Actor $j$ at time $t_2$ pushes the training data to create a new model denoted as $W_2$. However, at time $t_2$, Worker $j$’s model used in training (i.e. $W_0$) is stale compared to the target model (i.e. $W_1$). The stale update, shown as red arrows in Fig. \ref{fig:consist:sub1}, may slow down the convergence in training~\cite{Mnih2016,Li2018}. Distributed \textit{model-based} DRL methods also suffer from this issue. For example, \textit{reanalyze staleness} is incurred in SpeedyZero\cite{mei2023speedyzero}. This is because the parallel trainers commonly receive batches that have been reanalyzed by an outdated version of the model, rather than the latest target model.
%Besides, different workers with inconsistent model may also overwrite each other’s updates back and forth, thereby wasting computation and communication resources\cite{Mnih2016}.

\subsection{Synchronous On-policy training}

In order to mitigate the stale update issue mentioned above, the synchronous on-policy training method lets the parallel workers wait for the completion of gradient computation for all the workers, and then performs the model update synchronously. In this way, the models contained in parallel workers are consistent. This consistency among workers ensures the on-policy property of DRL algorithm that the behavior policy and the target policy are the same during training. Current synchronous on-policy training solutions in distributed DRL mainly use two ways to synchronize the model: centralized and decentralized manners.  For the former one such as PAAC~\cite{Clemente2017} and DBA3C~\cite{Adamski2018}, every worker transmits its gradients to the Parameter Server for model optimization. Then the updated model is copied to every worker. With respect to the decentralized manner such as DDPPO~\cite{Wijmans2019}, all worker’s gradients are shuffled among them. Then, the model in each worker is updated based on the aggregated gradients.  In both ways, the parallel workers update the model synchronously and consistently. It is clear in Fig. \ref{fig:consist:sub2} that there is no stale update that each worker holds the consistent model in each training iteration. 

However, it may incur the synchronization barrier, where the fast-running worker keeps idle and waits for the slow-running one in each iteration, as shown in Fig. \ref{fig:consist:sub2}. At time $t_1$, after Actor $i$ completes its training iteration, it waits for Actor $j$ to complete. This will deteriorate the utilization of computing resources in the cluster\cite{Wijmans2019}.
It is known that heterogeneity is the marked characteristic in the real cluster environment\cite{reiss2012heterogeneity}, where the case of processing at different speeds for different workers is quite common.  Besides, the gradient aggregation induces burst traffic while transmitting the gradients for all workers in a short duration. This may cause congestion and then reduce the utilization of communication resources.

\subsection{Discussion}

%In practical implementations, asynchronous off-policy training and synchronous on-policy training methods are both widely used in distributed DRL. 
Although asynchronous off-policy training methods have been shown higher resource utilizations than synchronous counterparts, they usually suffer from stability issues and converge to poorer results\cite{liu2020high,Pan2017}. Besides, recent works have demonstrated experimentally that synchronous on-policy training performs better than asynchronous off-policy training when using a large scale of distributed nodes. Adamski et al.~\cite{Adamski2018} witnessed that synchronous on-policy training outperformed asynchronous off-policy training in the experiments of training Atari games using 64 nodes.

To alleviate the synchronization barrier in synchronous training, stale-synchronous training introduces flexibility in the strict synchronization timing, allowing faster workers to proceed to the next iteration under the condition of bounded staleness (i.e. a concept of the maximum obsolete age between the slowest and other workers).  Once the staleness bounded is reached, a synchronous model update is enforced. In this way, a better trade-off between model consistency and resource utilization can be achieved. To the best of our knowledge, while stale-synchronous training achieves decent results in deep learning area\cite{zhao2019dynamic,Sun2021}, it is not yet widely adopted in distributed deep reinforcement learning currently. We believe stale-synchronous training holds great potential value in distributed deep reinforcement learning. One possible way is to enforce a synchronization according to the divergence between the target policy and the behavior policy.

Moreover, beyond vanilla on-policy and off-policy training, combining on-policy and off-policy in distributed DRL has attracted much attention in recent years.  Schmitt et al. \cite{schmitt2020off} propose to mix the off-policy replay experience with on-policy data and introduce a trust region algorithm that efficiently mitigates bias and enables efficient learning in distributed settings. Results show that this method outperforms IMPALA\cite{Espeholt2018} based on V-trace importance sampling. %Similar strategy has been utilized in \cite{banerjee2022improved}
Other than this, Borges et al. ~\cite{borges2021combining} propose to combine the off-policy targets with the on-policy targets in the distributed model-based DRL system named MuZero, improving the convergence speed and rewards.

\section{Deep Evolutionary Reinforcement Learning}\label{sec:ec}

%The dominant solution for training the neural network in DRL is backpropagation, which amends the parameters of the neural network based on stochastic gradient descent to maximize the expectation of the cumulative reward. On the other hand, evolutionary computation, an approach inspired by the process of natural selection, finds its great potential in solving DRL problems. In the interest of brevity, those evolutionary computation methods used in DRL are mainly in the field called neuroevolution~\cite{stanley2019designing} (also named Deep Evolutionary Reinforcement Learning in~\cite{Gupta,Khadka2019}). 
%, in which evolutionary algorithms are utilized to evolve neural networks. 
%Other than based on the backpropagation of gradients, neuroevolution methods directly perform searches in the parameter space by evolving a population of candidate solutions over many generations. As a result, neuroevolution methods enable massive parallelization if computing resources are sufficient. %Since neuroevolution methods do not rely on the Markov property and exploit the relationship between subsequent states, researchers argue that evolutionary methods are well-suited for problems with long time horizons, partial observation, and sparse reward structure~\cite{salimans2017evolution}. 

As mentioned above the dominant solution for training the neural network in distributed DRL is distributed stochastic gradient descent (DSGD) based on backpropagation of gradients. On the other hand, evolutionary computation, an approach inspired by the process of natural selection, finds its great potential in solving DRL problems. Many evolution-based training methods for distributed DRL have been proposed in recent years and have demonstrated impressive performance\cite{salimans2017evolution,tang2021guiding,Conti2018, such2017deep,Gupta,stanley2019designing}. This field of research is also named Deep Evolutionary Reinforcement Learning \cite{Gupta} and neuroevolution~\cite{stanley2019designing} in the literature.
%Other than using distributed stochastic gradient descent method that is based on backpropagation of gradients, an alternative approach, neuroevolution~\cite{stanley2019designing} (also named Deep Evolutionary Reinforcement Learning in~\cite{Gupta,Khadka2019}), finds its great potential in solving DRL problems. 
In the interest of brevity, evolution-based training methods directly perform searches in the parameter space by evolving a population of candidate solutions over many generations, without the need for backpropagating and aggregating gradients. As a result, evolution-based training methods enable massive parallelization with a low bandwidth requirement. 
%This field of research is also named Deep Evolutionary Reinforcement Learning \cite{Gupta} and neuroevolution~\cite{stanley2019designing} in the literature.

\begin{figure*}[!htb]
	\centering
	\subfigure[Evolution-guided Policy Gradient Method\cite{Khadka2019}]{
		\label{fig:ec:sub1} %% label for first subfigure
		\includegraphics[width=0.36\textwidth]{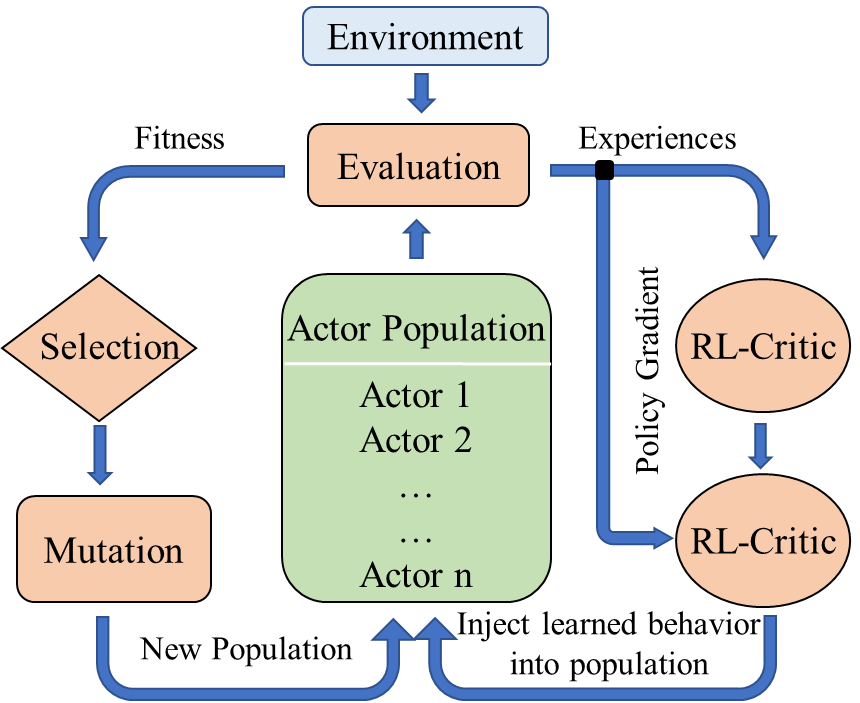}}
	\hspace{10mm}
	\subfigure[Deep Evolutionary Reinforcement Learning\cite{Gupta}]{
		\label{fig:ec:sub2} %% label for first subfigure
		\includegraphics[width=0.35\textwidth]{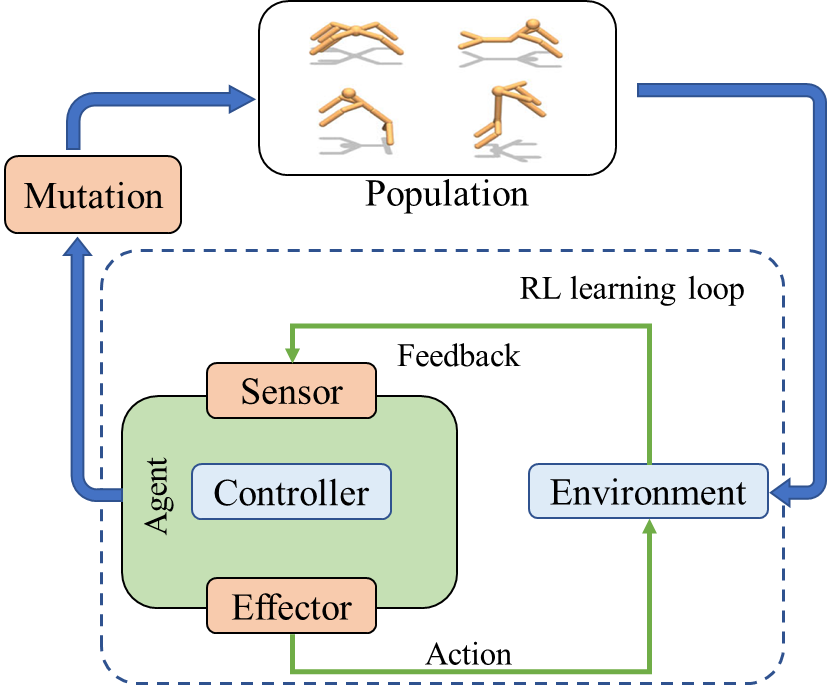}}
	\caption{Architectures of current methods based on a combination of learning and evolution.}
	\label{fig:ec}
\end{figure*}

\subsection{Training acceleration based on evolution strategies}
Evolution Strategies (ES) is one of the major branches of evolutionary computation, which iteratively updates a search distribution by using an estimated gradient on the parameter spaces~\cite{wierstra2014natural}.
There are many proposals that utilize evolution strategies algorithms to solve DRL problem~\cite{salimans2017evolution, Conti2018, heidrich2009uncertainty}. One representative research is~\cite{salimans2017evolution} proposed by Salimans et al. In this method, a population of parameters (\textit{genotypes}) and their objective function values (\textit{fitness}) for the neural network are maintained for every iteration (\textit{generation}). The parameters with the greatest fitness are selected to form the population for the next generation. This iteration is ended while the optimization objective is achieved. Let $F(\theta)$ represent the objective function parameterized by $\theta$, and $p_\mu$ denotes the distribution (parameterized by $\mu$) that the population follows. The goal of the method is to maximize the expectation value $\mathbb{E}_{\theta \sim p_\mu} F(\theta)$ over the population by searching for $\mu$ with stochastic gradient ascent. In terms of solving the DRL problem, the objective function $F(\theta)$ is the return after a sequence of actions $\textbf{a}$ is taken, denoted by $R(\textbf{a}(\theta))$, where the actions are determined by a policy $\textbf{a} = \pi(s|\theta)$. The method supposes the distribution $p_\mu$ as isotropic multivariate Gaussian with mean $\mu$ and fixed covariance $\sigma^2I$. Then, $\mathbb{E}_{\theta \sim p_\mu} F(\theta)$ can be written in the form of a Gaussian-blurred version of the original objective, that is,

\begin{equation}
	\mathbb{E}_{\theta \sim p_\mu}=\mathbb{E}_{\epsilon \sim \mathcal N(0,I)} F(\theta+\sigma\epsilon).
\end{equation}

Hence, the ES method performs the optimization by stochastic gradient ascent over $\theta$ directly with the following estimator:

\begin{equation}
	\nabla_\theta E_{\epsilon \sim \mathcal N(0,I)} F(\theta+\sigma\epsilon)=\frac{1}{\sigma} E_{\epsilon \sim \mathcal N(0,I)} F(\theta+\sigma\epsilon)\epsilon.
\end{equation}

Note that the ES method does not calculate gradients analytically but estimates the gradients of the objection function over parameters accordingly. By exploring in the parameter space instead of action space compared to policy gradient methods, the ES algorithms can be considered as a black box optimization that is invariant to action frequency and delayed rewards. This is thus better suited for problems with a long time horizon. Salimans et al.~\cite{salimans2017evolution} show that their algorithm achieves comparable results in continuous robotic control problems and video games. 

Another surprising advantage of the method in~\cite{salimans2017evolution} is the massive parallelization ability. This method incurs relatively low communication overhead when parallelized across many workers for the following reasons: First, since the operation in each iteration is conducted over the entire episode, the frequency of communication between workers is significantly reduced. Second, the information that needs to be transmitted between workers is limited to the return of an episode. This requires less bandwidth compared to distributed gradient-based methods, where the transfer involves gradients of the parameter vector or the parameter vector itself. Third, the elimination of value function approximations helps to reduce communication overhead, as there is no requirement for the additional gradient synchronization needed to learn the value function. Salimans et al.~\cite{salimans2017evolution} have deployed their algorithm over 80 machines with 1, 440 CPU cores, which achieves training time reduction by two orders of magnitude compared to single machine deployment. 

\subsection{Training acceleration based on genetic algorithms}
Genetic Algorithms (GA) is another branch of evolutionary computation, which is based on the theory about the genetic structure and behavior of chromosomes~\cite{kumar2010genetic}. Different from the ES-based method~\cite{salimans2017evolution} which is still gradient-based optimization, GA-based methods are gradient-free. Such et al.~\cite{such2017deep} propose a GA-based method called Deep GA for solving deep reinforcement-learning problem. Deep GA is based on a genetic algorithm which normally consists of three main operations: selection, mutation, crossover. In terms of selection, truncation selection is performed, which selects the top $T$ individuals as the parents of the next generation. The mutation is performed by adding Gaussian noise to the parameter vector shown as,
\begin{equation}
	\theta^n=\psi(\theta^{(n-1)},\tau_n)=\theta^{(n-1)}+\sigma\epsilon(\tau_n),
\end{equation}

\noindent where $\theta^n$ is the next generation of $\theta^{(n-1)}$, $\psi(\theta^{(n-1)},\tau_n)$ is a mutation function and $\tau_n$ is a list of mutation seeds for $\theta^n$. $\epsilon(\tau_n )$ follows Gaussian distribution $\mathcal N(0,I)$. With respect to crossover, Deep GA does not include it for simplicity. Besides, in order to scale well in the distributed setting, Deep GA leverages a compact encoding technique that uses an initialization seed plus a list of random seeds to reconstruct the parameter vector. The experiment results show that Deep GA can outperform ES\cite{salimans2017evolution}, A3C, and DQN on average. The results also reveal that Deep GA is superior to the gradient-based method in solving local optima problems by jumping across them in the parameter space.

Unlike only evolving weights of the neural networks in~\cite{such2017deep}, Stanley et al. propose a GA-based method called NEAT~\cite{stanley2002evolving,stanley2002efficient},  which evolves network topologies along with weights to enhance the learning efficiency. NEAT has been successfully applied to policy optimization and outperforms baseline with fixed-topology on a reinforcement learning benchmark. Although NEAT and HyperNEAT~\cite{stanley2009hypercube} represent early works for topology evolution of small networks, there are many recent works targeting deep (large-scale) neural networks~\cite{zoph2016neural, real2017large, real2019regularized,vargas2021review,Gupta}, including evolving the hyperparameters~\cite{baldominos2018evolutionary, zhang2021convolutional}. It is known that the architecture and hyperparameters of neural networks highly impact the overall performance. Success in the applications of deep (reinforcement) learning often requires fine-tuning of these building blocks of networks manually in practice. Fortunately, with the help of these solutions, the issue of manually designing a neural network for every application can be relieved.

\subsection{Hybridization of evolutionary computation and backpropagation}
The aforementioned methods mainly follow the path of performing a direct search in the parameter (or encodings) space based on the evolutionary computation, thereby obtaining the optimized neural networks that can serve as the policy in DRL problem. While evolutionary computation has merits of diverse exploration and massive parallelism based on population-based scheme, it suffers from high sample complexity, especially for problems with large numbers of parameters. Meanwhile, DRL methods leverage powerful backpropagation approaches to reinforce profitable actions into weight parameters. Hence, a hybrid scheme that combines the strengths of evolutionary computation and backpropagation has great potential value and triggers widely concerns. 

Khadka et al.~\cite{Khadka2019} propose an evolution-guided policy gradient method in DRL. This method incorporates evolutionary algorithms to generate diverse experiences and utilizes backpropagation in DRL to learn from them. More specifically, this method combines a genetic algorithm with DDPG algorithm, as shown in Fig. \ref{fig:ec:sub1}. A population of actor networks is maintained to interact with the environment and generate diverse experiences in parallel. The next generation of actors is created by selection, mutation, and crossover operators. Besides, the critic network and the actor network are trained by using backpropagation based on samples from the reply buffer. Then, the updated weights of the actor network are synchronized to the population of the actor networks.  In this way, this method injects the learned behaviors based on DRL into the evolving population. Experiments based on continuous control benchmarks show that this hybrid method outperforms prior DRL and evolutionary algorithms. 

Gangwani et al.~\cite{gangwani2018policy} propose a hybrid algorithm named Genetic Policy Optimization (GPO) for sample-efficient deep policy optimization. Rather than perform crossover in the parameter space in existing methods~\cite{stanley2002evolving,stanley2009hypercube,Khadka2019}, GPO does crossover in the state visitation space with the goal of cloning the behaviors of parents. Research showed that the crossover way of exchanging the parameters straightforwardly often leads to hierarchical relationship destruction and a loss of functionality~\cite{stanley2002evolving}. Besides, GPO utilizes the DRL algorithm (PPO) based on backpropagation instead of random permutation to mutate the weights of the actor networks. Experiments on Mojoco benchmark locomotion tasks demonstrate that GPO outperforms PPO and A2C\cite{baselines} algorithms and achieves comparable or higher sample efficiency.   

\subsection{Discussion}
Neuroevolution enables useful capabilities that are basically unavailable for traditional DRL approaches, including evolving building blocks of neural networks such as weights, topologies, and hyperparameters, facilitating diverse exploration and massive parallelization while maintaining the evolving population of networks. This also opens the door to incorporating learning and evolution to create sophisticated intelligence for solving high-dimensional decision-making problems. Other than the approaches above, there are still many interesting topics in this exciting area.  Some examples incorporate quality diversity~\cite{cully2015robots, pugh2016quality}, novelty search~\cite{conti2018improving, jackson2019novelty} and adaptive noise~\cite{plappert2017parameter, fortunato2017noisy} to improve exploration. Many works focus on Neural Architecture Search (NAS) by utilizing indirect encoding~\cite{fernandes2021pruning,Vargas-Hakim2022,assunccao2020incremental}, swarm intelligence algorithms~\cite{junior2019particle, wang2022dropout, darwish2020survey},  random search~\cite{liu2018darts, li2020random, zhang2020overcoming}, sequential model-based optimization~\cite{kandasamy2018neural, liu2018progressive}, etc. While NAS is still in the initial research stage, breakthroughs have been made in many field including image classification~\cite{simonyan2014very, zhu2020deep}, object detection~\cite{shih2019real, zhang2020discriminant}, machine translation~\cite{wu2016google, zhang2020neural}, multi-task learning~\cite{liang2018evolutionary, kim2021auto}. 

Besides, with the increasing capacities of computing resources that are easy to obtain, neuroevolution has further demonstrated its great potential in solving complex problems. Li et al. utilize deep evolutionary reinforcement learning to evolve diverse agent morphologies to handle locomotion and manipulation tasks in complex environments~\cite{Gupta}. Its architecture is shown in Fig. \ref{fig:ec:sub2}. This method employs 1152 CPUs to evolve ten generations of populations and train four thousand agent morphologies, with five million environment interactions for each morphology.  Real et al. propose to automatically discover neural network models for CIFAR-10 and CIFAR-100 datasets by evolving 1000 population of models with 250 parallel workers~\cite{real2017large}.  Asseman et al. study to use FPGA acceleration on Deep GA~\cite{such2017deep} method with a distributed system of 432 Xilinx FPGAs~\cite{Asseman2021}. In their experiment, 832 instances in parallel can be run, and 1.2 million frames per second of the aggregation rate can be achieved. Compared to the baseline method~\cite{such2017deep}, this method achieves twice speedup and high game scores in most of cases.

\section{Open-source Libraries and Platforms}\label{sec:lib}

In AI area that is flourishing, it is undoubted that developing and evaluating innovations in a rapid manner is essential in academia as well as the industrial community. This motivates researchers to develop tools, such as libraries and platforms, to facilitate DRL algorithms development in the field. In recent years, many libraries and platforms have been proposed in DRL areas, e.g., OpenAI baselines\footnote{https://github.com/openai/baselines}, Keras-RL\footnote{https://github.com/keras-rl/keras-rl}, MushroomRL\footnote{https://github.com/MushroomRL/mushroom-rl}, etc. Some of these are scalable and user-friendly that algorithmic components (e.g., functions, neural networks, environments, etc.) can be flexibly reused to compose customized learning agents. However, many of them are designed to be run on a single process or a single machine. Parallel and distributed execution at scale are not considered. Parallel and distributed programming have high professional restrictions that require complex operations, which are not friendly to developers who major in learning algorithm design. Therefore, there is a crucial demand to encapsulate parallelism primitives in libraries and platforms.

\begin{table*}[!htbp]
	\centering
	\caption{Open-source Libraries/Platforms for Parallel and Distributed Deep Reinforcement Learning. }
	{\scriptsize
		\begin{threeparttable}
			\renewcommand\arraystretch{1.5}
			\begin{tabular}{c|m{1.1cm}|c|m{3.4cm}|m{2.8cm}|m{4.0cm}}
				\toprule
				Libraries/Platforms &Institutes & Year & Baseline Algorithms & Supported Envs. & Parallel and Distributed Features \\
				\midrule
				TF-Agents\cite{TFAgents}	&Google	&2017 &DQN, DDQN, DDPG, TD3, REINFORCE, PPO, SAC, etc & Bsuite, DeepMind Control, Gym, MuJoCo, Pybullet &  Batch simulation and network forward pass parallelism based on CPUs/TPUs. \\
				
				\hline
				Ray RLlib\cite{liang2017ray}	&Berkeley 	&2017	&A2C, A3C, ARS, BC, DDPG, TD3, Rainbow, ES, APE-X, IMPALA, PPO, APPO, DD-PPO, SAC, QMIX, VDN, IQN, MADDPG, MCTS, etc.  & Gym, PettingZoo, Unity3D, Gazebo & %provides flexible APIs that abstract parallel programming details for users.
				 Synchronous training with straggler migrations, and  various baselines including evolution-based and model-based algorithms\\
				
				\hline
				ReAgent\cite{Gauci2019} 		&Facebook  &2018   	&DQN, Double DQN, Dueling DQN, QR-DQN, DDPG, TD3, SAC, etc. & Gym &  Workflows to perform training with large-scale data preprocessing, feature transformation, distributed training, etc.\\
				
				\hline
				SURREAL\cite{fan2018surreal} &Standford  &2018   &DDPG and PPO  & Gym, Deepmind Control, Surreal Robotics, MuJoCo & Asynchronous training and cluster automation setup on major cloud providers, e.g., AWS and Azure. \\
				
				\hline
				PARL\cite{parl} &Baidu	&2019 	&DQN, ES, DDPG, PPO, IMPALA, A2C, TD3, SAC, MADDPG, CQL, QMIX, etc.  & Gym & Parallelization of training with thousands of CPUs and multi-GPUs based on a centralized architecture. \\
				
				\hline
				rlpyt\cite{Stooke} &Berkeley  &2019 &A2C, PPO, DQN, DQN variants, Rainbow, R2D2, Ape-X, DDPG, TD3, SAC, etc. & Atari, Gym  & Synchronous and asynchronous sampling and optimization. \\
				
				\hline
				SEED\_RL\cite{espeholt2020seed} &Google  &2020 &IMPALA, R2D2, SAC, Vanilla PG, PPO, AWR, V-MPO, etc.   & Atari, DeepMind Control, Google Football, MuJoco  & Centralized inference and a fast communication layer based on gRPC.  \\
				
				\hline
				ACME\cite{Hoffman2020} &DeepMind  &2020  &D4PG, TD3, SAC, MPO, PPO, DMPO, MO-MPO, DQN, IMPALA, R2D2, MCTS etc.  & DeepMind Control, Gym &  Asynchronous training with a learner process and many distributed actor processes and low-level storage system \textit{Reverb} for experience replay. \\
				
				\hline
				Fiber\cite{Zhi2020} &Uber  &2020 &A3C, PPO, ES, etc. & ALE, Gym, MuJoCo  & Standard multiprocessing API and online migration from multiprocessing on one machine to cluster computing across multiple machines. \\
				
				\hline
				SampleFactory\cite{Petrenko2020} &Intel Lab  &2020 &PPO & MuJoCo, Atari, VizDoom, DeepMind Lab, Megaverse, Envpool, IsaacGym, Brax,
				Quad-Swarm-RL, Hugging Face Hub  &   Asynchronous training with off-policy corrections on a single-machine setting. \\
				
				\hline			
				ChainerRL\cite{JMLR:v22:20-376}	&Preferred Networks  &2021  & DQN, IQN, Rainbow, REINFORCE, A2C, ACER, PCL, DDPG, TRPO, PPO, TD3, SAC, etc. & Gym, MuJoCo  & Synchronous and asynchronous training. \\	
				
				\hline
				Tianshou\cite{tianshou}	& Tsinghua &2022  & REINFORCE, A2C, TRPO, PPO, DDPG, TD3, SAC, DQN, Rainbow, IQN, BCQ, CQL, CRR, PER, PSRL, etc.  & Gym, PettingZoo & Synchronous and asynchronous training, and standardization support of the training process.\\
				
%				\hline
%				Envpool\cite{weng2022envpool}	& Sea AI Lab &2022  & , etc.  & Atari, Mujoco, VizDoom, DeepMind Control Suite, etc. & \textcolor{blue}{Synchronous and asynchronous training}\\
				
				\hline
				TorchRL\cite{bou2023torchrl}	&UPF  &2023 &A2C, PPO, DDQN, SAC, REDQ, Dreamer, Decision transformers, RLHF, APPO, DPPO, DDPPO, IMPALA, Ape-X, etc.   & Gym, DeepMind Control  & Synchronous and asynchronous training, and high modularity that allows flexible composability for DRL training. \\
				
				\hline
				MindSpore\cite{zhu2023msrl}	&ICL, Huawei  &2023 &DQN, PPO, A2C, A3C, DDPG, MAPPO, MADDPG TD3, SAC, Double DQN, etc.  & Gym, MuJoCo, MPE, SMAC, DMC, PettingZoo, D4RL  &Synchronous and asynchronous training, and flexible parallelization using fragmented dataflow graph (FDG). \\
				
				\hline			
				SRL\cite{mei2023srl}	& Tsinghua  & 2024  & PPO  & Gym, Atari, Google Football, MuJoCo, Hide and Seek, SMAC & Synchronous training and data batch pre-fetching. \\
				
				\hline			
				 PEARL\cite{meng2024pearl}	&  Meta Research  &  2024  &  DQN, DDPG  &  Gym &  Portable implementations across heterogeneous platforms including FPGA and Optimization on DRL-specific runtime scheduling.\\
				\bottomrule
			\end{tabular}
			\begin{tablenotes}
				\footnotesize
				\item Statistics are collected from the Github links, documents, and papers of corresponding codebases in Aug. 2024.
			\end{tablenotes}
		\end{threeparttable}
	}\label{Tab.library}
\end{table*}

We review and compare current open-source libraries or platforms that support parallel and distributed DRL according to the following criteria.
\begin{itemize}
	\item Baseline Algorithms. They should provide rich choices of existing deep reinforcement learning algorithms, which facilitates rapid development and comparisons for users.
	\item Environment Integration. Simulation interaction is a basic element in DRL training. Hence, integrating simulation environments in training libraries and platforms plays an important role in supporting easy verification as well as applications. 
	\item Parallel and Distributed Features. The key parallel and distributed techniques and capabilities offered to enhance the DRL training efficiency are compared.  
	
	%\item Community. With new techniques continuously emerged in this rapid developing field, an active community that discusses the implementation details of promising approaches is demanding.
\end{itemize}

It is noted that we do not compare the performance of existing libraries and platforms experimentally, which is out of the scope of this survey article. Table \ref{Tab.library} provides an overview of the open-source libraries or platforms that support parallel and distributed DRL. Statistics are collected from the Github links, documents, and papers of corresponding codebases in Aug. 2024. 

%Due to the fact that DRL relies heavily on trail-on-error, using parallel and distributed computing to accelerate the training process becomes a fundamental need to tackle complex, realistic problems. Nevertheless, parallel and distributed programming have high professional restrictions that require complex operations, e.g., communication, synchronization, resource scheduling, etc. This is not friendly to developers who major in learning algorithm design. Therefore, there is a crucial demand to encapsulate parallelism primitives to support faster development in DRL. Table \ref{Tab.library} lists the open-source libraries or platforms that support distributed DRL.

Most of these libraries and platforms provide rich choices of baseline algorithms and supported environments. More specifically, RLlib~\cite{liang2017ray}, rlpyt~\cite{Stooke}, ACME~\cite{Hoffman2020} and TorchRL~\cite{bou2023torchrl} include value-based and policy gradient families of DRL algorithms, and provide baselines of the SOTA parallel and distributed DRL methods such as IMPALA~\cite{Espeholt2018} and Ape-X~\cite{Horgan2018}. In comparison, the open-source version of SampleFactory\cite{Petrenko2020} and SRL\cite{mei2023srl} only implement one algorithm (Proximal Policy Optimization, PPO). In terms of supported environments, all of the libraries and platforms integrate OpenAI's Gym and support Gym wrapper API. This provides a lot of convenience for users to access to environments. There are 9 codebases that support more than three types of environments, e.g., SampleFactory\cite{Petrenko2020} and MindSpore\cite{zhu2023msrl} provide 10 and 7 types of environments, respectively.

Besides, we compare the parallel and distributed features such as parallel training methods and communication mechanisms for different libraries and platforms.
RLlib~\cite{liang2017ray} and rlpyt~\cite{Stooke} offer training options of synchronous or asynchronous optimization. SEED\_RL~\cite{espeholt2020seed} encapsulates a fast communication layer based on streaming RPCs to mitigate the overhead of remote calls. Fiber~\cite{Zhi2020} provides standard multiprocessing API and supports migration from multiprocessing on one machine to cluster computing across multiple machines on the fly. Besides, some of them~\cite{fan2018surreal,liang2017ray} provide demonstrations or even automation cluster setups on major cloud providers, e.g., AWS, Azure, and Google Cloud. This is very useful to promote parallel and distributed DRL in practice since most research teams cannot afford to build a cluster of resources. %In additions, some of the libraries have very active communities. For instance, Ray RLlib  features a growing ecosystem of communities and has 28.1k Github stars and 886 contributors. The estimated response time for Github issues of Ray RLlib are less than 2 days. 

\section{Conclusion and Future Directions}\label{sec:conclusion}

Expanding DRL to a large-scale has become an inevitable trend nowadays. However, the rapid progress in which the field is developing makes it difficult to understand in a systematic manner. In this survey, we have provided a comprehensive literature review on training acceleration for DRL using parallel and distributed computing. In particular, we have analyzed the primary challenges to make DRL training distributed, and demystified the details of the technologies that have been proposed by researchers to address these challenges. This includes the system architectures for distributed DRL, the simulation parallelism to increase sample collection efficiency,  the computing parallelism to improve the computational efficiency, the distributed synchronization mechanisms for backpropagation-based training, the deep evolutionary reinforcement learning for evolution-based training.   Further, we have summarized the existing libraries and platforms in view of facilitating the research community in rapid development. This survey tries to enable readers to have a holistic view of distributed DRL and provides a guideline for them to get started quickly in this area. Below, we extrapolate potential directions for future work in this field.

%Deep reinforcement learning is constantly exploring the capability boundary of artificial intelligence; it is inevitable that the complexity of environments and tasks in its applications will have grown continuously. Besides, research towards Artificial General Intelligence (AGI) is in the ascendant, where a generalist DRL agent that can handle multi-task learning~\cite{Reed2022,zhang2021survey} and support knowledge transfer~\cite{d2020sharing,zhuang2020comprehensive} is desired. This drives AI to develop large models, such PromptPG~\cite{lu2022dynamic} and Q-learning Decision Transformer~\cite{yamagata2023q}, which creates new challenges and opportunities for parallel and distributed DRL algorithms.
%Techniques such as pipeline parallelism \cite{liang2023survey} and model compression \cite{deng2020model} that can decompose large networks are worthy of exploring in DRL. Further, manually designing the architecture for a large model  may go beyond human strength. Automated neural architecture search \cite{elsken2019neural} would be a desirable direction for future work.  

%\textbf{In-memory computing.} 

\textbf{Specialized hardware accelerators.} Designing specialized hardware accelerators that are computationally efficient with neural networks becomes an urgent and promising direction. It has been witnessed that many domain-specific architectures have been proposed to accelerate DRL training such as NVIDIA Tensor Core\cite{Stooke2018}, FPGAs\cite{Cho2019}, TPUs\cite{espeholt2020seed}. %Recent works have shown that TPUs significantly improve the computational efficiency compared to GPUs\cite{wang2019benchmarking,wang2020benchmarking}. 
Despite the advancements in computation speed, several emerging topics remain in this highly dynamic field of research. One such direction is in-memory computing, which is critical for reducing data movement and minimizing latency for DRL training\cite{sebastian2020memory,Meng2022}. Another direction is the exploitation of pipeline parallelism according to the DRL training workloads, which requires innovative hardware designs and algorithms to ensure all the arrays on the hardware are always active during training \cite{wang2020fpdeep}.  Additionally, how to 
reduce the energy consumption while improving the computational efficiency is also very important. We believe that neuromorphic hardware design is a promising direction, such as spike-based neuromorphic chips~\cite{rao2022long}. Moreover, heterogeneous architectures that combine the strengths of different hardware platforms in the field of DRL can be further investigated\cite{meng2024pearl}.  

%although specialized hardware accelerators are good at batch processing workloads, there are many other works such as scheduling, sampling, data I/O that cannot saturate their computation power.
%heterogeneous architectures, enabling efficient interplay of CPUs, GPUs, and FPGAs in the domain of Reinforcement Learning, should be further investigated.

%Additionally, the interconnection of multiple nodes in a distributed computing environment poses a significant challenge, particularly in terms of achieving high bandwidth and low latency communication.

\textbf{In-network distributed aggregation mechanisms.} Network communication occupies a large part of the execution time in distributed gradient aggregation. To alleviate the communication overhead, an emerging trend is to shift the gradient aggregation process from the worker nodes to the network infrastructure itself, such as at the programmable switches. By doing so, the gradient aggregation is conducted on-the-fly at the granularity of network packets rather than gradient vectors stored within the memory of the worker nodes. This approach has the potential to substantially reduce the end-to-end network latency and the volume of data transferred across the network during distributed training. Li et al. \cite{Li2019} were pioneers in leveraging in-switch computing to accelerate the distributed training of deep reinforcement learning, achieving significant results. We believe designing packet forwarding and computing mechanisms according to the characteristics of DRL workloads will have great potential for training improvement.

\textbf{Efficient sample exploitation algorithms.} A large amount of rollout data is required is one of the known issues of DRL training. One reason is counted for the low sample reuse rate. OpenAI researchers demonstrate that the sample reuse rate is lower than 1 in the asynchronous training of the OpenAI Five\cite{berner2019dota}. Note that simply collecting more data in parallel without effectively exploiting it does not necessarily translate into improved training outcomes. %Actually, rollout samples for off-policy learning can be used many times\cite{de2018experience}. 
Therefore, the development of algorithms that can exploit rollout samples more effectively, without compromising the stability of the learning process, is a crucial direction for future research in distributed DRL. To address this challenge, several strategies can be pursued. First, more sophisticated replay mechanisms, including Priority-refresh\cite{mei2023speedyzero}, asynchronous curriculum experience replay\cite{hu2023asynchronous}, and regret minimization replay\cite{liu2021regret}, are promising techniques to better utilize the samples. Besides, driving DRL agents to perform searches sophisticatedly rather than randomly is helping for generating more useful samples. Exploration strategies such as curiosity-driven~\cite{schwarke2023curiosity}, diversity-driven~\cite{wu2022quality}, and novelty search \cite{stanley2019designing} in large state-action spaces are potential directions of importance.

\textbf{Large language model-enhanced DRL.} Large language models (LLMs), equipped with extensive pre-trained knowledge and powerful generalization abilities, are emerging as a promising direction to enhance the performance of DRL, including accelerating the training process. Specifically, in the DRL training paradigm, LLMs can assist DRL agents in reward function design, action selection, and policy evaluation, based on the modeling capability and common-sense pre-trained knowledge. In this way, LLMs not only enable a considerable level of ability at the beginning of the training process, but also facilitate the decision-making during the optimization. Currently, numerous pioneering works \cite{bai2022training,chu2023accelerating,cao2024survey} have been proposed that I am confident will pave the way for robust development in the coming future. 

%\textbf{High-throughput stale-synchronization mechanisms.} As mentioned in Section~\ref{sec:sync} that the stale-synchronous training mechanism, which relaxes the strict synchronization timing, can achieve a better trade-off between model consistency and the resource utilization. However, there are few studies exploring stale-synchronous training in distributed DRL.  

%Programmable switches have been widely used in clusters nowadays, which provides computation capacities for gradient aggregation in the network. 
%Li et al. ~\cite{Li2019} propose an in-switch computing method to accelerate distributed reinforcement learning, which moves the gradient aggregation from server nodes to the programmable network switches to relieve the communication burden. This method can speed-up by more than 3 times compared with state-of-the-art distributed training methods. Similar finding is illustrated in \cite{gebara2021network,sapio2019scaling}, where in-network aggregation is proposed. 

%Thirdly, network communication occupies a large part of the execution time in each training iteration. Hence, efficient communication architectures are desired as the scale of the training data increases. 
%Besides, gradient compression techniques including quantization \cite{markov2023quantized,sun2019communication} and sparsification \cite{zhang2022mipd,chen2020scalecom} draw much attention in reducing data transfer workloads during distributed training. However, since less information is gathered, this kind of methods may scarify the model convergence performance. 

\section{Acknowledgments}
This work was supported by the National Natural Science Foundation of China under grant U21A20518, 62025208 and 62421002.

\bibliographystyle{unsrt}
\bibliography{uav}

\begin{thebibliography}{100}

\bibitem{mnih2015human}
Volodymyr Mnih, Koray Kavukcuoglu, David Silver, Andrei~A Rusu, Joel Veness,
  Marc~G Bellemare, Alex Graves, Martin Riedmiller, Andreas~K Fidjeland, Georg
  Ostrovski, et~al.
\newblock Human-level control through deep reinforcement learning.
\newblock {\em Nature}, 518(7540):529--533, 2015.

\bibitem{Silver2016}
David Silver, Aja Huang, Chris~J. Maddison, Arthur Guez, Laurent Sifre, George
  {Van Den Driessche}, Julian Schrittwieser, Ioannis Antonoglou, Veda
  Panneershelvam, Marc Lanctot, Sander Dieleman, Dominik Grewe, John Nham, Nal
  Kalchbrenner, Ilya Sutskever, Timothy Lillicrap, Madeleine Leach, Koray
  Kavukcuoglu, Thore Graepel, and Demis Hassabis.
\newblock {Mastering the game of Go with deep neural networks and tree search}.
\newblock {\em Nature}, 529(7587):484--489, 2016.

\bibitem{Silver2017}
David Silver, Thomas Hubert, Julian Schrittwieser, Ioannis Antonoglou, Matthew
  Lai, Arthur Guez, Marc Lanctot, Laurent Sifre, Dharshan Kumaran, Thore
  Graepel, et~al.
\newblock A general reinforcement learning algorithm that masters chess, shogi,
  and go through self-play.
\newblock {\em Science}, 362(6419):1140--1144, 2018.

\bibitem{vinyals2019grandmaster}
Oriol Vinyals, Igor Babuschkin, Wojciech~M Czarnecki, Micha{\"{e}}l Mathieu,
  Andrew Dudzik, Junyoung Chung, David~H Choi, Richard Powell, Timo Ewalds,
  Petko Georgiev, and Others.
\newblock {Grandmaster level in StarCraft II using multi-agent reinforcement
  learning}.
\newblock {\em Nature}, 575(7782):350--354, 2019.

\bibitem{kaufmann2023champion}
Elia Kaufmann, Leonard Bauersfeld, Antonio Loquercio, Matthias M{\"u}ller,
  Vladlen Koltun, and Davide Scaramuzza.
\newblock Champion-level drone racing using deep reinforcement learning.
\newblock {\em Nature}, 620(7976):982--987, 2023.

\bibitem{Andrychowicz2020}
Open AI:~Marcin Andrychowicz, Bowen Baker, Maciek Chociej, Rafal
  J{\'{o}}zefowicz, Bob McGrew, Jakub Pachocki, Arthur Petron, Matthias
  Plappert, Glenn Powell, Alex Ray, Jonas Schneider, Szymon Sidor, Josh Tobin,
  Peter Welinder, Lilian Weng, and Wojciech Zaremba.
\newblock {Learning dexterous in-hand manipulation}.
\newblock {\em International Journal of Robotics Research}, 39(1):3--20, 2020.

\bibitem{fan2018surreal}
Linxi Fan, Yuke Zhu, Jiren Zhu, Zihua Liu, Orien Zeng, Anchit Gupta, Joan
  Creus-Costa, Silvio Savarese, and Li~Fei-Fei.
\newblock Surreal: Open-source reinforcement learning framework and robot
  manipulation benchmark.
\newblock In {\em Conference on Robot Learning}, pages 767--782. PMLR, 2018.

\bibitem{Gupta}
Agrim Gupta, Silvio Savarese, Surya Ganguli, and Li~Fei-fei.
\newblock {Embodied intelligence via learning and evolution}.
\newblock {\em Nature Communications}, (2021).

\bibitem{popova2018deep}
Mariya Popova, Olexandr Isayev, and Alexander Tropsha.
\newblock Deep reinforcement learning for de novo drug design.
\newblock {\em Science advances}, 4(7):eaap7885, 2018.

\bibitem{yala2022optimizing}
Adam Yala, Peter~G Mikhael, Constance Lehman, Gigin Lin, Fredrik Strand,
  Yung-Liang Wan, Kevin Hughes, Siddharth Satuluru, Thomas Kim, Imon Banerjee,
  et~al.
\newblock Optimizing risk-based breast cancer screening policies with
  reinforcement learning.
\newblock {\em Nature Medicine}, 28(1):136--143, 2022.

\bibitem{saria2018individualized}
Suchi Saria.
\newblock Individualized sepsis treatment using reinforcement learning.
\newblock {\em Nature medicine}, 24(11):1641--1642, 2018.

\bibitem{henderson2018deep}
Peter Henderson, Riashat Islam, Philip Bachman, Joelle Pineau, Doina Precup,
  and David Meger.
\newblock Deep reinforcement learning that matters.
\newblock In {\em Proceedings of the AAAI conference on artificial
  intelligence}, volume~32, 2018.

\bibitem{Arulkumaran2017}
Kai Arulkumaran, Marc~Peter Deisenroth, Miles Brundage, and Anil~Anthony
  Bharath.
\newblock {Deep reinforcement learning: A brief survey}.
\newblock {\em IEEE Signal Processing Magazine}, 34(6):26--38, 2017.

\bibitem{li2017deep}
Yuxi Li.
\newblock Deep reinforcement learning: An overview.
\newblock {\em arXiv preprint arXiv:1701.07274}, 2017.

\bibitem{9904958}
Xu~Wang, Sen Wang, Xingxing Liang, Dawei Zhao, Jincai Huang, Xin Xu, Bin Dai,
  and Qiguang Miao.
\newblock Deep reinforcement learning: A survey.
\newblock {\em IEEE Transactions on Neural Networks and Learning Systems},
  pages 1--15, 2022.

\bibitem{moerland2023model}
Thomas~M Moerland, Joost Broekens, Aske Plaat, Catholijn~M Jonker, et~al.
\newblock Model-based reinforcement learning: A survey.
\newblock {\em Foundations and Trends{\textregistered} in Machine Learning},
  16(1):1--118, 2023.

\bibitem{Kiran2022}
B.~Ravi Kiran, Ibrahim Sobh, Victor Talpaert, Patrick Mannion, Ahmad~A.Al
  Sallab, Senthil Yogamani, and Patrick Perez.
\newblock {Deep Reinforcement Learning for Autonomous Driving: A Survey}.
\newblock {\em IEEE Transactions on Intelligent Transportation Systems},
  23(6):4909--4926, 2022.

\bibitem{Nguyen2021a}
Thanh~Thi Nguyen and Vijay~Janapa Reddi.
\newblock {Deep Reinforcement Learning for Cyber Security}.
\newblock {\em IEEE Transactions on Neural Networks and Learning Systems},
  (Ml), 2021.

\bibitem{Luong2019}
Nguyen~Cong Luong, Dinh~Thai Hoang, Shimin Gong, Dusit Niyato, Ping Wang,
  Ying~Chang Liang, and Dong~In Kim.
\newblock {Applications of Deep Reinforcement Learning in Communications and
  Networking: A Survey}.
\newblock {\em IEEE Communications Surveys and Tutorials}, 21(4):3133--3174,
  2019.

\bibitem{Haydari2022}
Ammar Haydari and Yasin Yilmaz.
\newblock {Deep Reinforcement Learning for Intelligent Transportation Systems:
  A Survey}.
\newblock {\em IEEE Transactions on Intelligent Transportation Systems},
  23(1):11--32, 2022.

\bibitem{Nguyen2020}
Thanh~Thi Nguyen, Ngoc~Duy Nguyen, and Saeid Nahavandi.
\newblock {Deep Reinforcement Learning for Multiagent Systems: A Review of
  Challenges, Solutions, and Applications}.
\newblock {\em IEEE Transactions on Cybernetics}, 50(9):3826--3839, 2020.

\bibitem{chen2021deep}
Wuhui Chen, Xiaoyu Qiu, Ting Cai, Hong-Ning Dai, Zibin Zheng, and Yan Zhang.
\newblock Deep reinforcement learning for internet of things: A comprehensive
  survey.
\newblock {\em IEEE Communications Surveys \& Tutorials}, 23(3):1659--1692,
  2021.

\bibitem{verbraeken2020survey}
Joost Verbraeken, Matthijs Wolting, Jonathan Katzy, Jeroen Kloppenburg, Tim
  Verbelen, and Jan~S Rellermeyer.
\newblock A survey on distributed machine learning.
\newblock {\em Acm computing surveys (csur)}, 53(2):1--33, 2020.

\bibitem{peteiro2013survey}
Diego Peteiro-Barral and Bertha Guijarro-Berdi{\~n}as.
\newblock A survey of methods for distributed machine learning.
\newblock {\em Progress in Artificial Intelligence}, 2:1--11, 2013.

\bibitem{wang2020survey}
Meng Wang, Weijie Fu, Xiangnan He, Shijie Hao, and Xindong Wu.
\newblock A survey on large-scale machine learning.
\newblock {\em IEEE Transactions on Knowledge and Data Engineering},
  34(6):2574--2594, 2020.

\bibitem{Mayer2020}
Ruben Mayer and Hans~Arno Jacobsen.
\newblock {Scalable deep learning on distributed infrastructures: Challenges,
  techniques, and tools}.
\newblock {\em ACM Computing Surveys}, 53(1), 2020.

\bibitem{Ben-Nun2019}
Tal Ben-Nun and Torsten Hoefler.
\newblock {Demystifying parallel and distributed deep learning: An in-depth
  concurrency analysis}.
\newblock {\em ACM Computing Surveys}, 52(4), 2019.

\bibitem{Chahal2020}
Karanbir~Singh Chahal, Manraj~Singh Grover, Kuntal Dey, and Rajiv~Ratn Shah.
\newblock {A Hitchhiker's Guide On Distributed Training Of Deep Neural
  Networks}.
\newblock {\em Journal of Parallel and Distributed Computing}, 137:65--76,
  2020.

\bibitem{Samsami2020}
Mohammad~Reza Samsami and Hossein Alimadad.
\newblock {Distributed Deep Reinforcement Learning: An Overview}.
\newblock {\em arXiv preprint arXiv:2011.11012}, pages 1--15, 2020.

\bibitem{yin2024distributed}
Qiyue Yin, Tongtong Yu, Shengqi Shen, Jun Yang, Meijing Zhao, Wancheng Ni,
  Kaiqi Huang, Bin Liang, and Liang Wang.
\newblock Distributed deep reinforcement learning: A survey and a multi-player
  multi-agent learning toolbox.
\newblock {\em Machine Intelligence Research}, 21(3):411--430, 2024.

\bibitem{clavera2018model}
Ignasi Clavera, Jonas Rothfuss, John Schulman, Yasuhiro Fujita, Tamim Asfour,
  and Pieter Abbeel.
\newblock Model-based reinforcement learning via meta-policy optimization.
\newblock In {\em Conference on Robot Learning}, pages 617--629. PMLR, 2018.

\bibitem{kurutach2018model}
Thanard Kurutach, Ignasi Clavera, Yan Duan, Aviv Tamar, and Pieter Abbeel.
\newblock Model-ensemble trust-region policy optimization.
\newblock {\em arXiv preprint arXiv:1802.10592}, 2018.

\bibitem{sutton2018reinforcement}
Richard~S Sutton.
\newblock Reinforcement learning: an introduction.
\newblock {\em A Bradford Book}, 2018.

\bibitem{schulman2017proximal}
John Schulman, Filip Wolski, Prafulla Dhariwal, Alec Radford, and Oleg Klimov.
\newblock Proximal policy optimization algorithms.
\newblock {\em arXiv preprint arXiv:1707.06347}, 2017.

\bibitem{haarnoja2018soft}
Tuomas Haarnoja, Aurick Zhou, Pieter Abbeel, and Sergey Levine.
\newblock Soft actor-critic: Off-policy maximum entropy deep reinforcement
  learning with a stochastic actor.
\newblock In {\em International Conference on Machine Learning}, pages
  1861--1870. PMLR, 2018.

\bibitem{fujimoto2018addressing}
Scott Fujimoto, Herke Hoof, and David Meger.
\newblock Addressing function approximation error in actor-critic methods.
\newblock In {\em International Conference on Machine Learning}, pages
  1587--1596. PMLR, 2018.

\bibitem{fu2019diagnosing}
Justin Fu, Aviral Kumar, Matthew Soh, and Sergey Levine.
\newblock Diagnosing bottlenecks in deep q-learning algorithms.
\newblock In {\em International Conference on Machine Learning}, pages
  2021--2030. PMLR, 2019.

\bibitem{kumar2019stabilizing}
Aviral Kumar, Justin Fu, Matthew Soh, George Tucker, and Sergey Levine.
\newblock Stabilizing off-policy q-learning via bootstrapping error reduction.
\newblock In {\em NeurIPS}, volume~32, 2019.

\bibitem{chu2006map}
Cheng-Tao Chu, Sang Kim, Yi-An Lin, YuanYuan Yu, Gary Bradski, Kunle Olukotun,
  and Andrew Ng.
\newblock Map-reduce for machine learning on multicore.
\newblock In {\em NeurIPS}, volume~19, 2006.

\bibitem{zinkevich2010parallelized}
Martin Zinkevich, Markus Weimer, Lihong Li, and Alex Smola.
\newblock Parallelized stochastic gradient descent.
\newblock In {\em NeurIPS}, volume~23, 2010.

\bibitem{NIPS2014_1ff1de77}
Mu~Li, David~G Andersen, Alexander~J Smola, and Kai Yu.
\newblock Communication efficient distributed machine learning with the
  parameter server.
\newblock In Z.~Ghahramani, M.~Welling, C.~Cortes, N.~Lawrence, and K.Q.
  Weinberger, editors, {\em NeurIPS}, volume~27. Curran Associates, Inc., 2014.

\bibitem{10.1145/3035918.3035933}
Jiawei Jiang, Bin Cui, Ce~Zhang, and Lele Yu.
\newblock Heterogeneity-aware distributed parameter servers.
\newblock In {\em Proceedings of the ACM SIGMOD}, page 463–478, New York, NY,
  USA, 2017. Association for Computing Machinery.

\bibitem{zhang2020autosync}
Hao Zhang, Yuan Li, Zhijie Deng, Xiaodan Liang, Lawrence Carin, and Eric Xing.
\newblock Autosync: Learning to synchronize for data-parallel distributed deep
  learning.
\newblock In {\em NeurIPS}, volume~33, pages 906--917, 2020.

\bibitem{dean2012large}
Jeffrey Dean, Greg Corrado, Rajat Monga, Kai Chen, Matthieu Devin, Mark Mao,
  Marc'aurelio Ranzato, Andrew Senior, Paul Tucker, Ke~Yang, et~al.
\newblock Large scale distributed deep networks.
\newblock In {\em NeurIPS}, volume~25, 2012.

\bibitem{lee2014model}
Seunghak Lee, Jin~Kyu Kim, Xun Zheng, Qirong Ho, Garth~A Gibson, and Eric~P
  Xing.
\newblock On model parallelization and scheduling strategies for distributed
  machine learning.
\newblock In {\em NeurIPS}, volume~27, 2014.

\bibitem{jain2020gems}
Arpan Jain, Ammar~Ahmad Awan, Asmaa~M Aljuhani, Jahanzeb~Maqbool Hashmi,
  Quentin~G Anthony, Hari Subramoni, Dhableswar~K Panda, Raghu Machiraju, and
  Anil Parwani.
\newblock Gems: Gpu-enabled memory-aware model-parallelism system for
  distributed dnn training.
\newblock In {\em SC20: International Conference for High Performance
  Computing, Networking, Storage and Analysis}, pages 1--15. IEEE, 2020.

\bibitem{narayanan2021efficient}
Deepak Narayanan, Mohammad Shoeybi, Jared Casper, Patrick LeGresley, Mostofa
  Patwary, Vijay Korthikanti, Dmitri Vainbrand, Prethvi Kashinkunti, Julie
  Bernauer, Bryan Catanzaro, et~al.
\newblock Efficient large-scale language model training on gpu clusters using
  megatron-lm.
\newblock In {\em Proceedings of the International Conference for High
  Performance Computing, Networking, Storage and Analysis}, pages 1--15, 2021.

\bibitem{Tang2020}
Zhenheng Tang, Shaohuai Shi, Xiaowen Chu, Wei Wang, and Bo~Li.
\newblock {Communication-Efficient Distributed Deep Learning: A Comprehensive
  Survey}.
\newblock {\em arXiv preprint arXiv:2003.06307}, (1):1--23, 2020.

\bibitem{Huang2019}
Yanping Huang, Youlong Cheng, Ankur Bapna, Orhan Firat, Mia~Xu Chen, Dehao
  Chen, Hyouk~Joong Lee, Jiquan Ngiam, Quoc~V. Le, Yonghui Wu, and Zhifeng
  Chen.
\newblock {GPipe: Efficient training of giant neural networks using pipeline
  parallelism}.
\newblock In {\em NeurIPS}, volume~32, 2019.

\bibitem{Narayanan2019}
Deepak Narayanan, Aaron Harlap, Amar Phanishayee, Vivek Seshadri, Nikhil~R.
  Devanur, Gregory~R. Ganger, Phillip~B. Gibbons, and Matei Zaharia.
\newblock {Pipedream: Generalized pipeline parallelism for DNN training}.
\newblock {\em SOSP 2019 - Proceedings of the 27th ACM Symposium on Operating
  Systems Principles}, pages 1--15, 2019.

\bibitem{narayanan2021memory}
Deepak Narayanan, Amar Phanishayee, Kaiyu Shi, Xie Chen, and Matei Zaharia.
\newblock Memory-efficient pipeline-parallel dnn training.
\newblock In {\em International Conference on Machine Learning}, pages
  7937--7947. PMLR, 2021.

\bibitem{Luo2022}
Ziyue Luo, Xiaodong Yi, Guoping Long, Shiqing Fan, Chuan Wu, Jun Yang, and Wei
  Lin.
\newblock {Efficient Pipeline Planning for Expedited Distributed DNN Training}.
\newblock (Ml):340--349, 2022.

\bibitem{Krizhevsky2014}
Alex Krizhevsky.
\newblock {One weird trick for parallelizing convolutional neural networks}.
\newblock {\em arXiv preprint arXiv:1404.5997}, 2014.

\bibitem{Wu2016}
Yonghui Wu, Mike Schuster, Zhifeng Chen, Quoc~V. Le, Mohammad Norouzi, Wolfgang
  Macherey, Maxim Krikun, Yuan Cao, Qin Gao, Klaus Macherey, Jeff Klingner,
  Apurva Shah, Melvin Johnson, Xiaobing Liu, {\L}ukasz Kaiser, Stephan Gouws,
  Yoshikiyo Kato, Taku Kudo, Hideto Kazawa, Keith Stevens, George Kurian,
  Nishant Patil, Wei Wang, Cliff Young, Jason Smith, Jason Riesa, Alex Rudnick,
  Oriol Vinyals, Greg Corrado, Macduff Hughes, and Jeffrey Dean.
\newblock {Google's Neural Machine Translation System: Bridging the Gap between
  Human and Machine Translation}.
\newblock {\em arXiv preprint arXiv:1609.08144}, pages 1--23, 2016.

\bibitem{NEURIPS2018_3a37abde}
Noam Shazeer, Youlong Cheng, Niki Parmar, Dustin Tran, Ashish Vaswani, Penporn
  Koanantakool, Peter Hawkins, HyoukJoong Lee, Mingsheng Hong, Cliff Young,
  Ryan Sepassi, and Blake Hechtman.
\newblock Mesh-tensorflow: Deep learning for supercomputers.
\newblock In S.~Bengio, H.~Wallach, H.~Larochelle, K.~Grauman, N.~Cesa-Bianchi,
  and R.~Garnett, editors, {\em NeurIPS}, volume~31. Curran Associates, Inc.,
  2018.

\bibitem{Jia2019}
Zhihao Jia, Matei Zaharia, and Alex Aiken.
\newblock {BEYOND DATA AND MODEL PARALLELISM FOR DEEP NEURAL NETWORKS}.
\newblock 2019.

\bibitem{Rajbhandari2020}
Samyam Rajbhandari, Jeff Rasley, Olatunji Ruwase, and Yuxiong He.
\newblock {Zero: Memory optimizations toward training trillion parameter
  models}.
\newblock {\em International Conference for High Performance Computing,
  Networking, Storage and Analysis, SC}, 2020-November:3505--3506, 2020.

\bibitem{park2020hetpipe}
Jay~H Park, Gyeongchan Yun, M~Yi Chang, Nguyen~T Nguyen, Seungmin Lee, Jaesik
  Choi, Sam~H Noh, and Young-ri Choi.
\newblock $\{$HetPipe$\}$: Enabling large $\{$DNN$\}$ training on (whimpy)
  heterogeneous $\{$GPU$\}$ clusters through integration of pipelined model
  parallelism and data parallelism.
\newblock In {\em 2020 USENIX Annual Technical Conference (USENIX ATC 20)},
  pages 307--321, 2020.

\bibitem{Bian2021}
Zhengda Bian, Hongxin Liu, Boxiang Wang, Haichen Huang, Yongbin Li, Chuanrui
  Wang, Fan Cui, and Yang You.
\newblock {Colossal-AI: A Unified Deep Learning System For Large-Scale Parallel
  Training}.
\newblock {\em arXiv preprint arXiv:2110.14883}, 2021.

\bibitem{Petrenko2020}
Aleksei Petrenko, Zhehui Huang, Tushar Kumar, Gaurav Sukhatme, and Vladlen
  Koltun.
\newblock {Sample factory: Egocentric 3D control from pixels at 100000 FPS with
  asynchronous reinforcement learning}.
\newblock In {\em International Conference on Machine Learning}, volume
  PartF16814, pages 7608--7618. PMLR, 2020.

\bibitem{Espeholt2018}
Lasse Espeholt, Hubert Soyer, Remi Munos, Karen Simonyan, Volodymyr Mnih, Tom
  Ward, Boron Yotam, Firoiu Vlad, Harley Tim, Iain Dunning, Shane Legg, and
  Koray Kavukcuoglu.
\newblock {IMPALA: Scalable Distributed Deep-RL with Importance Weighted
  Actor-Learner Architectures}.
\newblock In {\em International Conference on Machine Learning}, volume~4,
  pages 2263--2284. PMLR, 2018.

\bibitem{Makoviychuk2021}
Viktor Makoviychuk, Lukasz Wawrzyniak, Yunrong Guo, Michelle Lu, Kier Storey,
  Miles Macklin, David Hoeller, Nikita Rudin, Arthur Allshire, Ankur Handa, and
  Gavriel State.
\newblock {Isaac Gym: High Performance GPU-Based Physics Simulation For Robot
  Learning}.
\newblock {\em arXiv preprint arXiv:2108.10470}, 2021.

\bibitem{baker2020emergent}
Bowen Baker, Ingmar Kanitscheider, Todor Markov, Yi~Wu, Glenn Powell, Bob
  McGrew, and Igor Mordatch.
\newblock Emergent tool use from multi-agent autocurricula.
\newblock In {\em International Conference on Learning Representations}, 2020.

\bibitem{machupalli2022review}
Raju Machupalli, Masum Hossain, and Mrinal Mandal.
\newblock Review of asic accelerators for deep neural network.
\newblock {\em Microprocessors and Microsystems}, 89:104441, 2022.

\bibitem{Chen2021a}
Tianyi Chen, Kaiqing Zhang, Georgios~B. Giannakis, and Tamer Basar.
\newblock {Communication-Efficient Policy Gradient Methods for Distributed
  Reinforcement Learning}.
\newblock {\em IEEE Transactions on Control of Network Systems}, pages 1--14,
  2021.

\bibitem{such2017deep}
Felipe~Petroski Such, Vashisht Madhavan, Edoardo Conti, Joel Lehman, Kenneth~O
  Stanley, and Jeff Clune.
\newblock Deep neuroevolution: Genetic algorithms are a competitive alternative
  for training deep neural networks for reinforcement learning.
\newblock {\em arXiv preprint arXiv:1712.06567}, 2017.

\bibitem{Horgan2018}
Dan Horgan, John Quan, David Budden, Gabriel Barth-Maron, Matteo Hessel, Hado
  {Van Hasselt}, and David Silver.
\newblock {Distributed prioritized experience replay}.
\newblock In {\em International Conference on Learning Representations}, pages
  1--19, 2018.

\bibitem{kapturowski2018recurrent}
Steven Kapturowski, Georg Ostrovski, John Quan, Remi Munos, and Will Dabney.
\newblock Recurrent experience replay in distributed reinforcement learning.
\newblock In {\em International Conference on Learning Representations}, 2018.

\bibitem{Nair2015}
Arun Nair, Praveen Srinivasan, Sam Blackwell, Cagdas Alcicek, Rory Fearon,
  Alessandro {De Maria}, Vedavyas Panneershelvam, Mustafa Suleyman, Charles
  Beattie, Stig Petersen, Shane Legg, Volodymyr Mnih, Koray Kavukcuoglu, and
  David Silver.
\newblock {Massively Parallel Methods for Deep Reinforcement Learning}.
\newblock {\em arXiv preprint arXiv:1507.04296}, 2015.

\bibitem{Mnih2016}
Volodymyr Mnih, Adria~Puigdomenech Badia, Mehdi Mirza, Alex Graves, Timothy
  Lillicrap, Tim Harley, David Silver, and Koray Kavukcuoglu.
\newblock Asynchronous methods for deep reinforcement learning.
\newblock In {\em International Conference on Machine Learning}, pages
  1928--1937. PMLR, 2016.

\bibitem{Stooke}
Adam Stooke and Pieter Abbeel.
\newblock {rlpyt : A Research Code Base for Deep Reinforcement Learning in
  PyTorch}.
\newblock pages 1--12.

\bibitem{Wijmans2019}
Erik Wijmans, Abhishek Kadian, Ari Morcos, Stefan Lee, Irfan Essa, Devi Parikh,
  Manolis Savva, and Dhruv Batra.
\newblock {DD-PPO: Learning Near-Perfect PointGoal Navigators from 2.5 Billion
  Frames}.
\newblock In {\em International Conference on Learning Representations}, pages
  1--21, 2020.

\bibitem{liang2018rllib}
Eric Liang, Richard Liaw, Robert Nishihara, Philipp Moritz, Roy Fox, Ken
  Goldberg, Joseph Gonzalez, Michael Jordan, and Ion Stoica.
\newblock Rllib: Abstractions for distributed reinforcement learning.
\newblock In {\em International Conference on Machine Learning}, pages
  3053--3062. PMLR, 2018.

\bibitem{Li2019}
Youjie Li, Iou~Jen Liu, Yifan Yuan, Deming Chen, Alexander Schwing, and Jian
  Huang.
\newblock {Accelerating distributed reinforcement learning with in-switch
  computing}.
\newblock {\em Proceedings - International Symposium on Computer Architecture},
  pages 279--291, 2019.

\bibitem{Stooke2018}
Adam Stooke and Pieter Abbeel.
\newblock {Accelerated Methods for Deep Reinforcement Learning}.
\newblock {\em arXiv preprint arXiv:1803.02811}, 2018.

\bibitem{yazdanbakhsh2020menger}
Amir Yazdanbakhsh, Junchao Chen, and Yu~Zheng.
\newblock Menger: Massively large-scale distributed reinforcement learning.
\newblock 2020.

\bibitem{espeholt2020seed}
Lasse Espeholt, Rapha{\"e}l Marinier, Piotr Stanczyk, Ke~Wang, and Marcin
  Michalski.
\newblock Seed rl: Scalable and efficient deep-rl with accelerated central
  inference.
\newblock In {\em International Conference on Learning Representations}, 2020.

\bibitem{mei2023srl}
Zhiyu Mei, Wei Fu, Guangju Wang, Huanchen Zhang, and Yi~Wu.
\newblock Srl: Scaling distributed reinforcement learning to over ten thousand
  cores.
\newblock In {\em International Conference on Learning Representations}, 2024.

\bibitem{li2023parallel}
Zechu Li, Tao Chen, Zhang-Wei Hong, Anurag Ajay, and Pulkit Agrawal.
\newblock Parallel $ q $-learning: Scaling off-policy reinforcement learning
  under massively parallel simulation.
\newblock In {\em International Conference on Machine Learning}, pages
  19440--19459. PMLR, 2023.

\bibitem{zhang2020asynchronous}
Yunzhi Zhang, Ignasi Clavera, Boren Tsai, and Pieter Abbeel.
\newblock Asynchronous methods for model-based reinforcement learning.
\newblock In {\em Conference on Robot Learning}, pages 1338--1347. PMLR, 2020.

\bibitem{assran2019gossip}
Mahmoud Assran, Joshua Romoff, Nicolas Ballas, Joelle Pineau, and Michael
  Rabbat.
\newblock Gossip-based actor-learner architectures for deep reinforcement
  learning.
\newblock In {\em NeurIPS}, volume~32, 2019.

\bibitem{Mathkar2017}
Adwaitvedant Mathkar and Vivek~S. Borkar.
\newblock {Distributed Reinforcement Learning via Gossip}.
\newblock {\em IEEE Transactions on Automatic Control}, 62(3):1465--1470, 2017.

\bibitem{sha2022fully}
Xingyu Sha, Jiaqi Zhang, Keyou You, Kaiqing Zhang, and Tamer Ba{\c{s}}ar.
\newblock Fully asynchronous policy evaluation in distributed reinforcement
  learning over networks.
\newblock {\em Automatica}, 136:110092, 2022.

\bibitem{Hessel2018}
Matteo Hessel, Joseph Modayil, Hado {Van Hasselt}, Tom Schaul, Georg Ostrovski,
  Will Dabney, Dan Horgan, Bilal Piot, Mohammad Azar, and David Silver.
\newblock {Rainbow: Combining improvements in deep reinforcement learning}.
\newblock {\em 32nd AAAI Conference on Artificial Intelligence, AAAI 2018},
  pages 3215--3222, 2018.

\bibitem{Gruslys2018}
Audrūnas Gruslys, Will Dabney, Mohammad~Gheshlaghi Azar, Bilal Piot, Marc~G.
  Bellemare, and R{\'{e}}mi Munos.
\newblock {The reactor: A fast and sample-efficient actor-critic agent for
  reinforcement learning}.
\newblock In {\em International Conference on Learning Representations}, pages
  1--18, 2018.

\bibitem{schulman2015trust}
John Schulman, Sergey Levine, Pieter Abbeel, Michael Jordan, and Philipp
  Moritz.
\newblock Trust region policy optimization.
\newblock In {\em International Conference on Machine Learning}, pages
  1889--1897. PMLR, 2015.

\bibitem{lillicrap2015continuous}
Timothy~P Lillicrap, Jonathan~J Hunt, Alexander Pritzel, Nicolas Heess, Tom
  Erez, Yuval Tassa, David Silver, and Daan Wierstra.
\newblock Continuous control with deep reinforcement learning.
\newblock {\em arXiv preprint arXiv:1509.02971}, 2015.

\bibitem{Schulman2017}
John Schulman, Filip Wolski, Prafulla Dhariwal, Alec Radford, and Oleg Klimov.
\newblock {Proximal Policy Optimization Algorithms}.
\newblock {\em arXiv preprint arXiv:1707.06347}, pages 1--12, 2017.

\bibitem{Haarnoja2018}
Tuomas Haarnoja, Aurick Zhou, Kristian Hartikainen, George Tucker, Sehoon Ha,
  Jie Tan, Vikash Kumar, Henry Zhu, Abhishek Gupta, Pieter Abbeel, and Sergey
  Levine.
\newblock {Soft Actor-Critic Algorithms and Applications}.
\newblock {\em arXiv preprint arXiv:1812.05905}, 2018.

\bibitem{Tian2017}
Yuandong Tian, Qucheng Gong, Wenling Shang, Yuxin Wu, and C.~Lawrence Zitnick.
\newblock {ELF: An extensive, lightweight and flexible research platform for
  real-time strategy games}.
\newblock In {\em NeurIPS}, number Nips, pages 2660--2670, 2017.

\bibitem{freeman2021brax}
C~Daniel Freeman, Erik Frey, Anton Raichuk, Sertan Girgin, Igor Mordatch, and
  Olivier Bachem.
\newblock Brax--a differentiable physics engine for large scale rigid body
  simulation.
\newblock {\em arXiv preprint arXiv:2106.13281}, 2021.

\bibitem{mittal2015timely}
Radhika Mittal, Vinh~The Lam, Nandita Dukkipati, Emily Blem, Hassan Wassel,
  Monia Ghobadi, Amin Vahdat, Yaogong Wang, David Wetherall, and David Zats.
\newblock Timely: Rtt-based congestion control for the datacenter.
\newblock {\em ACM SIGCOMM Computer Communication Review}, 45(4):537--550,
  2015.

\bibitem{liu2016dynamic}
Zhihong Liu, Qi~Zhang, Reaz Ahmed, Raouf Boutaba, Yaping Liu, and Zhenghu Gong.
\newblock Dynamic resource allocation for mapreduce with partitioning skew.
\newblock {\em IEEE Transactions on Computers}, 65(11):3304--3317, 2016.

\bibitem{Dalton2020}
Steven Dalton and Iuri Frosio.
\newblock {Accelerating reinforcement learning through GPU Atari emulation}.
\newblock In {\em NeurIPS}, 2020.

\bibitem{guo2019limits}
Jing Guo, Zihao Chang, Sa~Wang, Haiyang Ding, Yihui Feng, Liang Mao, and
  Yungang Bao.
\newblock Who limits the resource efficiency of my datacenter: An analysis of
  alibaba datacenter traces.
\newblock In {\em Proceedings of the International Symposium on Quality of
  Service}, pages 1--10, 2019.

\bibitem{Liang2018}
Jacky Liang, Viktor Makoviychuk, Ankur Handa, Nuttapong Chentanez, Miles
  Macklin, and Dieter Fox.
\newblock {GPU-Accelerated Robotic Simulation for Distributed Reinforcement
  Learning}.
\newblock {\em arXiv preprint arXiv:1810.05762}, (CoRL):1--14, 2018.

\bibitem{salimans2017evolution}
Tim Salimans, Jonathan Ho, Xi~Chen, Szymon Sidor, and Ilya Sutskever.
\newblock Evolution strategies as a scalable alternative to reinforcement
  learning.
\newblock {\em arXiv preprint arXiv:1703.03864}, 2017.

\bibitem{shacklett2021large}
Brennan Shacklett, Erik Wijmans, Aleksei Petrenko, Manolis Savva, Dhruv Batra,
  Vladlen Koltun, and Kayvon Fatahalian.
\newblock Large batch simulation for deep reinforcement learning.
\newblock In {\em International Conference on Learning Representations}, 2021.

\bibitem{peng2021amp}
Xue~Bin Peng, Ze~Ma, Pieter Abbeel, Sergey Levine, and Angjoo Kanazawa.
\newblock Amp: Adversarial motion priors for stylized physics-based character
  control.
\newblock {\em ACM Transactions on Graphics (TOG)}, 40(4):1--20, 2021.

\bibitem{Freeman2021}
C.~Daniel Freeman, Erik Frey, Anton Raichuk, Sertan Girgin, Igor Mordatch, and
  Olivier Bachem.
\newblock {Brax -- A Differentiable Physics Engine for Large Scale Rigid Body
  Simulation}.
\newblock {\em arXiv preprint arXiv:2106.13281}, 2021.

\bibitem{rudin22a}
Nikita Rudin, David Hoeller, Philipp Reist, and Marco Hutter.
\newblock Learning to walk in minutes using massively parallel deep
  reinforcement learning.
\newblock In Aleksandra Faust, David Hsu, and Gerhard Neumann, editors, {\em
  Proceedings of the 5th Conference on Robot Learning}, volume 164 of {\em
  Proceedings of Machine Learning Research}, pages 91--100. PMLR, 08--11 Nov
  2022.

\bibitem{loquercio2021learning}
Antonio Loquercio, Elia Kaufmann, Ren{\'e} Ranftl, Matthias M{\"u}ller, Vladlen
  Koltun, and Davide Scaramuzza.
\newblock Learning high-speed flight in the wild.
\newblock {\em Science Robotics}, 6(59):eabg5810, 2021.

\bibitem{dean2008mapreduce}
Jeffrey Dean and Sanjay Ghemawat.
\newblock Mapreduce: simplified data processing on large clusters.
\newblock {\em Communications of the ACM}, 51(1):107--113, 2008.

\bibitem{li2011mapreduce}
Yuxi Li and Dale Schuurmans.
\newblock Mapreduce for parallel reinforcement learning.
\newblock In {\em European Workshop on Reinforcement Learning}, pages 309--320.
  Springer, 2011.

\bibitem{liang2017ray}
Eric Liang, Richard Liaw, Robert Nishihara, Philipp Moritz, Roy Fox, Joseph
  Gonzalez, Ken Goldberg, and Ion Stoica.
\newblock Ray rllib: A composable and scalable reinforcement learning library.
\newblock {\em arXiv preprint arXiv:1712.09381}, 85, 2017.

\bibitem{Mania2018}
Horia Mania, Aurelia Guy, and Benjamin Recht.
\newblock {Simple random search provides a competitive approach to
  reinforcement learning}.
\newblock {\em arXiv preprint arXiv:1803.07055}, pages 1--22, 2018.

\bibitem{Adamski2018}
Igor Adamski, Robert Adamski, Tomasz Grel, Adam J{\k{e}}drych, Kamil Kaczmarek,
  and Henryk Michalewski.
\newblock Distributed deep reinforcement learning: Learn how to play atari
  games in 21 minutes.
\newblock In {\em ISC High Performance 2018, Frankfurt, Germany}, pages
  370--388. Springer, 2018.

\bibitem{Heess2017}
Nicolas Heess, Dhruva TB, Srinivasan Sriram, Jay Lemmon, Josh Merel, Greg
  Wayne, Yuval Tassa, Tom Erez, Ziyu Wang, S.~M.~Ali Eslami, Martin Riedmiller,
  and David Silver.
\newblock {Emergence of Locomotion Behaviours in Rich Environments}.
\newblock {\em arXiv preprint arXiv:1707.02286}, 2017.

\bibitem{Radients2018}
P~Olicy~G Radients, Gabriel Barth-maron, Matthew~W Hoffman, David Budden, Will
  Dabney, Dan Horgan, Dhruva Tb, Alistair Muldal, Nicolas Heess, and Timothy
  Lillicrap.
\newblock Distributed distributional deterministic policy gradients.
\newblock In {\em International Conference on Learning Representations}, pages
  1--16.

\bibitem{Babaeizadeh2017}
Mohammad Babaeizadeh, Iuri Frosio, Stephen Tyree, Jason Clemons, and Jan Kautz.
\newblock {Reinforcement learning through asynchronous advantage actor-critic
  on a GPU}.
\newblock In {\em International Conference on Learning Representations}, pages
  1--12, 2017.

\bibitem{clemente2017efficient}
Alfredo~V Clemente, Humberto~N Castej{\'o}n, and Arjun Chandra.
\newblock Efficient parallel methods for deep reinforcement learning.
\newblock {\em arXiv preprint arXiv:1705.04862}, 2017.

\bibitem{zhu2023msrl}
Huanzhou Zhu, Bo~Zhao, Gang Chen, Weifeng Chen, Yijie Chen, Liang Shi, Yaodong
  Yang, Peter Pietzuch, and Lei Chen.
\newblock $\{$MSRL$\}$: Distributed reinforcement learning with dataflow
  fragments.
\newblock In {\em 2023 USENIX Annual Technical Conference (USENIX ATC 23)},
  pages 977--993, 2023.

\bibitem{berner2019dota}
Christopher Berner, Greg Brockman, Brooke Chan, Vicki Cheung, Przemys{\l}aw
  D{\k{e}}biak, Christy Dennison, David Farhi, Quirin Fischer, Shariq Hashme,
  Chris Hesse, et~al.
\newblock Dota 2 with large scale deep reinforcement learning.
\newblock {\em arXiv preprint arXiv:1912.06680}, 2019.

\bibitem{mei2023speedyzero}
Yixuan Mei, Jiaxuan Gao, Weirui Ye, Shaohuai Liu, Yang Gao, and Yi~Wu.
\newblock Speedyzero: Mastering atari with limited data and time.
\newblock In {\em International Conference on Learning Representations}, 2023.

\bibitem{Su2017}
Jiang Su, Jianxiong Liu, David~B. Thomas, and Peter~Y.K. Cheung.
\newblock {Neural Network Based Reinforcement Learning Acceleration on FPGA
  Platforms}.
\newblock {\em ACM SIGARCH Computer Architecture News}, 44(4):68--73, 2017.

\bibitem{Shao2017}
Shengjia Shao and Wayne Luk.
\newblock {Customised pearlmutter propagation: A hardware architecture for
  trust region policy optimisation}.
\newblock {\em 2017 27th International Conference on Field Programmable Logic
  and Applications, FPL 2017}, 2017.

\bibitem{Guo2019}
Ce~Guo, Wayne Luk, Stanley Qing~Shui Loh, Alexander Warren, and Joshua Levine.
\newblock {Customisable control policy learning for robotics}.
\newblock {\em Proceedings of the International Conference on
  Application-Specific Systems, Architectures and Processors},
  2019-July:91--98, 2019.

\bibitem{Cho2019}
Hyungmin Cho, Pyeongseok Oh, Jiyoung Park, Wookeun Jung, and Jaejin Lee.
\newblock {FA3C: FPGA-Accelerated Deep Reinforcement Learning}.
\newblock {\em International Conference on Architectural Support for
  Programming Languages and Operating Systems - ASPLOS}, pages 499--513, 2019.

\bibitem{Meng2020}
Yuan Meng, Sanmukh Kuppannagari, and Viktor Prasanna.
\newblock {Accelerating Proximal Policy Optimization on CPU-FPGA Heterogeneous
  Platforms}.
\newblock {\em Proceedings - 28th IEEE International Symposium on
  Field-Programmable Custom Computing Machines, FCCM 2020}, pages 19--27, 2020.

\bibitem{Meng2022}
Yuan Meng, Chi Zhang, and Viktor Prasanna.
\newblock {FPGA acceleration of deep reinforcement learning using on-chip
  replay management}.
\newblock pages 40--48, 2022.

\bibitem{Reed2022}
Scott Reed, Konrad Zolna, Emilio Parisotto, Sergio~Gomez Colmenarejo, Alexander
  Novikov, Gabriel Barth-Maron, Mai Gimenez, Yury Sulsky, Jackie Kay,
  Jost~Tobias Springenberg, Tom Eccles, Jake Bruce, Ali Razavi, Ashley Edwards,
  Nicolas Heess, Yutian Chen, Raia Hadsell, Oriol Vinyals, Mahyar Bordbar, and
  Nando de~Freitas.
\newblock {A Generalist Agent}.
\newblock {\em arXiv preprint arXiv:2205.06175}, pages 1--40, 2022.

\bibitem{Zhi2020}
Jiale Zhi, Rui Wang, Jeff Clune, and Kenneth~O. Stanley.
\newblock {Fiber: A Platform for Efficient Development and Distributed Training
  for Reinforcement Learning and Population-Based Methods}.
\newblock {\em arXiv preprint arXiv:2003.11164}, (Oliphant 2006), 2020.

\bibitem{da2018parallel}
Lucileide~MD Da~Silva, Matheus~F Torquato, and Marcelo~AC Fernandes.
\newblock Parallel implementation of reinforcement learning q-learning
  technique for fpga.
\newblock {\em IEEE Access}, 7:2782--2798, 2018.

\bibitem{rajat2020qtaccel}
Rachit Rajat, Yuan Meng, Sanmukh Kuppannagari, Ajitesh Srivastava, Viktor
  Prasanna, and Rajgopal Kannan.
\newblock Qtaccel: A generic fpga based design for q-table based reinforcement
  learning accelerators.
\newblock In {\em Proceedings of the 2020 ACM/SIGDA International Symposium on
  Field-Programmable Gate Arrays}, pages 323--323, 2020.

\bibitem{Ray18}
Philipp Moritz, Robert Nishihara, Stephanie Wang, Alexey Tumanov, Richard Liaw,
  Eric Liang, Melih Elibol, Zongheng Yang, William Paul, Michael~I. Jordan, and
  Ion Stoica.
\newblock Ray: a distributed framework for emerging ai applications.
\newblock In {\em OSDI'18: Proceedings of the 13th USENIX conference on
  Operating Systems Design and Implementation}, 2018.

\bibitem{yu2024cheaper}
Hanfei Yu, Jian Li, Yang Hua, Xu~Yuan, and Hao Wang.
\newblock Cheaper and faster: Distributed deep reinforcement learning with
  serverless computing.
\newblock In {\em Proceedings of the AAAI Conference on Artificial
  Intelligence}, volume~38, pages 16539--16547, 2024.

\bibitem{meng2024pearl}
Yuan Meng, Michael Kinsner, Deshanand Singh, Mahesh Iyer, and Viktor Prasanna.
\newblock Pearl: Enabling portable, productive, and high-performance deep
  reinforcement learning using heterogeneous platforms.
\newblock In {\em Proceedings of the 21st ACM International Conference on
  Computing Frontiers}, pages 41--50, 2024.

\bibitem{Bou2020}
Albert Bou and Gianni {De Fabritiis}.
\newblock {PyTorchRL: Modular and Distributed Reinforcement Learning in
  PyTorch}.
\newblock {\em arXiv preprint arXiv:2007.02622}, 2020.

\bibitem{Pan2017}
Xinghao Pan, Jianmin Chen, Rajat Monga, Samy Bengio, and Rafal Jozefowicz.
\newblock {Revisiting Distributed Synchronous SGD}.
\newblock {\em arXiv preprint arXiv:1702.05800}, pages 1--10, 2017.

\bibitem{Lian2018}
Xiangru Lian, Wei Zhang, Ce~Zhang, and Ji~Liu.
\newblock {Asynchronous decentralized parallel stochastic gradient descent}.
\newblock In {\em International Conference on Machine Learning}, volume~7,
  pages 4745--4767. PMLR, 2018.

\bibitem{Ho2013}
Qirong Ho, James Cipar, Henggang Cui, Jin~Kyu Kim, Seunghak Lee, Phillip~B.
  Gibbons, Garth~A. Gibson, Gregory~R. Ganger, and Eric~P. Xing.
\newblock {More effective distributed ML via a stale synchronous parallel
  parameter server}.
\newblock In {\em NeurIPS}, number~1, pages 1--9, 2013.

\bibitem{Li2018}
Youjie Li, Mingchao Yu, Songze Li, Salman Avestimehr, Nam~Sung Kim, and
  Alexander Schwing.
\newblock {PIPE-SGD: A decentralized pipelined SGD framework for distributed
  deep net training}.
\newblock In {\em NeurIPS}, volume 2018-Decem, pages 8045--8056, 2018.

\bibitem{Clemente2017}
Alfredo~V. Clemente, Humberto~N. Castej{\'{o}}n, and Arjun Chandra.
\newblock {Efficient Parallel Methods for Deep Reinforcement Learning}.
\newblock {\em arXiv preprint arXiv:1705.04862}, pages 1--9, 2017.

\bibitem{reiss2012heterogeneity}
Charles Reiss, Alexey Tumanov, Gregory~R Ganger, Randy~H Katz, and Michael~A
  Kozuch.
\newblock Heterogeneity and dynamicity of clouds at scale: Google trace
  analysis.
\newblock In {\em Proceedings of the third ACM symposium on cloud computing},
  pages 1--13, 2012.

\bibitem{liu2020high}
Iou-Jen Liu, Raymond Yeh, and Alexander Schwing.
\newblock High-throughput synchronous deep rl.
\newblock {\em Advances in Neural Information Processing Systems},
  33:17070--17080, 2020.

\bibitem{zhao2019dynamic}
Xing Zhao, Aijun An, Junfeng Liu, and Bao~Xin Chen.
\newblock Dynamic stale synchronous parallel distributed training for deep
  learning.
\newblock In {\em 2019 IEEE 39th International Conference on Distributed
  Computing Systems (ICDCS)}, pages 1507--1517. IEEE, 2019.

\bibitem{Sun2021}
Haifeng Sun, Zhiyi Gui, Song Guo, Qi~Qi, Jingyu Wang, and Jianxi Liao.
\newblock {GSSP: Eliminating Stragglers through Grouping Synchronous for
  Distributed Deep Learning in Heterogeneous Cluster}.
\newblock {\em IEEE Transactions on Cloud Computing}, 14(8), 2021.

\bibitem{schmitt2020off}
Simon Schmitt, Matteo Hessel, and Karen Simonyan.
\newblock Off-policy actor-critic with shared experience replay.
\newblock In {\em International Conference on Machine Learning}, pages
  8545--8554. PMLR, 2020.

\bibitem{borges2021combining}
Alexandre Borges and Arlindo Oliveira.
\newblock Combining off and on-policy training in model-based reinforcement
  learning.
\newblock {\em arXiv preprint arXiv:2102.12194}, 2021.

\bibitem{tang2021guiding}
Yunhao Tang.
\newblock Guiding evolutionary strategies with off-policy actor-critic.
\newblock In {\em Proceedings of the 20th International Conference on
  Autonomous Agents and MultiAgent Systems}, pages 1317--1325, 2021.

\bibitem{Conti2018}
Edoardo Conti, Joel Lehman, and Kenneth~O Stanley.
\newblock {Improving Exploration in Evolution Strategies for Deep Reinforcement
  Learning via a Population of Novelty-Seeking Agents}.
\newblock (NeurIPS), 2018.

\bibitem{stanley2019designing}
Kenneth~O Stanley, Jeff Clune, Joel Lehman, and Risto Miikkulainen.
\newblock Designing neural networks through neuroevolution.
\newblock {\em Nature Machine Intelligence}, 1(1):24--35, 2019.

\bibitem{Khadka2019}
Shauharda Khadka.
\newblock {Evolution-Guided Policy Gradient in Reinforcement Learning}.
\newblock (November), 2019.

\bibitem{wierstra2014natural}
Daan Wierstra, Tom Schaul, Tobias Glasmachers, Yi~Sun, Jan Peters, and
  J{\"u}rgen Schmidhuber.
\newblock Natural evolution strategies.
\newblock {\em The Journal of Machine Learning Research}, 15(1):949--980, 2014.

\bibitem{heidrich2009uncertainty}
Verena Heidrich-Meisner and Christian Igel.
\newblock Uncertainty handling cma-es for reinforcement learning.
\newblock In {\em Proceedings of the 11th Annual conference on Genetic and
  evolutionary computation}, pages 1211--1218, 2009.

\bibitem{kumar2010genetic}
Manoj Kumar, Dr~Husain, Naveen Upreti, Deepti Gupta, et~al.
\newblock Genetic algorithm: Review and application.
\newblock {\em Available at SSRN 3529843}, 2010.

\bibitem{stanley2002evolving}
Kenneth~O Stanley and Risto Miikkulainen.
\newblock Evolving neural networks through augmenting topologies.
\newblock {\em Evolutionary computation}, 10(2):99--127, 2002.

\bibitem{stanley2002efficient}
Kenneth~O Stanley and Risto Miikkulainen.
\newblock Efficient reinforcement learning through evolving neural network
  topologies.
\newblock In {\em Proceedings of the 4th Annual Conference on genetic and
  evolutionary computation}, pages 569--577, 2002.

\bibitem{stanley2009hypercube}
Kenneth~O Stanley, David~B D'Ambrosio, and Jason Gauci.
\newblock A hypercube-based encoding for evolving large-scale neural networks.
\newblock {\em Artificial life}, 15(2):185--212, 2009.

\bibitem{zoph2016neural}
Barret Zoph and Quoc~V Le.
\newblock Neural architecture search with reinforcement learning.
\newblock In {\em International Conference on Learning Representations}, 2017.

\bibitem{real2017large}
Esteban Real, Sherry Moore, Andrew Selle, Saurabh Saxena, Yutaka~Leon Suematsu,
  Jie Tan, Quoc~V Le, and Alexey Kurakin.
\newblock Large-scale evolution of image classifiers.
\newblock In {\em International Conference on Machine Learning}, pages
  2902--2911. PMLR, 2017.

\bibitem{real2019regularized}
Esteban Real, Alok Aggarwal, Yanping Huang, and Quoc~V Le.
\newblock Regularized evolution for image classifier architecture search.
\newblock In {\em Proceedings of the aaai conference on artificial
  intelligence}, volume~33, pages 4780--4789, 2019.

\bibitem{vargas2021review}
Gustavo-Adolfo Vargas-Hakim, Efren Mezura-Montes, and Hector-Gabriel
  Acosta-Mesa.
\newblock A review on convolutional neural network encodings for
  neuroevolution.
\newblock {\em IEEE Transactions on Evolutionary Computation}, 26(1):12--27,
  2021.

\bibitem{baldominos2018evolutionary}
Alejandro Baldominos, Yago Saez, and Pedro Isasi.
\newblock Evolutionary convolutional neural networks: An application to
  handwriting recognition.
\newblock {\em Neurocomputing}, 283:38--52, 2018.

\bibitem{zhang2021convolutional}
Miao Zhang, Huiqi Li, Shirui Pan, Juan Lyu, Steve Ling, and Steven Su.
\newblock Convolutional neural networks-based lung nodule classification: A
  surrogate-assisted evolutionary algorithm for hyperparameter optimization.
\newblock {\em IEEE Transactions on Evolutionary Computation}, 25(5):869--882,
  2021.

\bibitem{gangwani2018policy}
Tanmay Gangwani and Jian Peng.
\newblock Policy optimization by genetic distillation.
\newblock In {\em International Conference on Learning Representations}, 2018.

\bibitem{baselines}
Prafulla Dhariwal, Christopher Hesse, Oleg Klimov, Alex Nichol, Matthias
  Plappert, Alec Radford, John Schulman, Szymon Sidor, Yuhuai Wu, and Peter
  Zhokhov.
\newblock Openai baselines.
\newblock \url{https://github.com/openai/baselines}, 2017.

\bibitem{cully2015robots}
Antoine Cully, Jeff Clune, Danesh Tarapore, and Jean-Baptiste Mouret.
\newblock Robots that can adapt like animals.
\newblock {\em Nature}, 521(7553):503--507, 2015.

\bibitem{pugh2016quality}
Justin~K Pugh, Lisa~B Soros, and Kenneth~O Stanley.
\newblock Quality diversity: A new frontier for evolutionary computation.
\newblock {\em Frontiers in Robotics and AI}, page~40, 2016.

\bibitem{conti2018improving}
Edoardo Conti, Vashisht Madhavan, Felipe Petroski~Such, Joel Lehman, Kenneth
  Stanley, and Jeff Clune.
\newblock Improving exploration in evolution strategies for deep reinforcement
  learning via a population of novelty-seeking agents.
\newblock In {\em NeurIPS}, volume~31, 2018.

\bibitem{jackson2019novelty}
Ethan~C Jackson and Mark Daley.
\newblock Novelty search for deep reinforcement learning policy network weights
  by action sequence edit metric distance.
\newblock In {\em Proceedings of the Genetic and Evolutionary Computation
  Conference Companion}, pages 173--174, 2019.

\bibitem{plappert2017parameter}
Matthias Plappert, Rein Houthooft, Prafulla Dhariwal, Szymon Sidor, Richard~Y
  Chen, Xi~Chen, Tamim Asfour, Pieter Abbeel, and Marcin Andrychowicz.
\newblock Parameter space noise for exploration.
\newblock In {\em International Conference on Learning Representations}, 2018.

\bibitem{fortunato2017noisy}
Meire Fortunato, Mohammad~Gheshlaghi Azar, Bilal Piot, Jacob Menick, Ian
  Osband, Alex Graves, Vlad Mnih, Remi Munos, Demis Hassabis, Olivier Pietquin,
  et~al.
\newblock Noisy networks for exploration.
\newblock In {\em International Conference on Learning Representations}, 2018.

\bibitem{fernandes2021pruning}
Francisco~E Fernandes~Jr and Gary~G Yen.
\newblock Pruning deep convolutional neural networks architectures with
  evolution strategy.
\newblock {\em Information Sciences}, 552:29--47, 2021.

\bibitem{Vargas-Hakim2022}
Gustavo~Adolfo Vargas-H{\'{a}}kim, Efr{\'{e}}n Mezura-Montes, and
  H{\'{e}}ctor~Gabriel Acosta-Mesa.
\newblock {A Review on Convolutional Neural Network Encodings for
  Neuroevolution}.
\newblock {\em IEEE Transactions on Evolutionary Computation}, 26(1):12--27,
  2022.

\bibitem{assunccao2020incremental}
Filipe Assun{\c{c}}{\~a}o, Nuno Louren{\c{c}}o, Bernardete Ribeiro, and
  Penousal Machado.
\newblock Incremental evolution and development of deep artificial neural
  networks.
\newblock In {\em European Conference on Genetic Programming (Part of
  EvoStar)}, pages 35--51. Springer, 2020.

\bibitem{junior2019particle}
Francisco Erivaldo~Fernandes Junior and Gary~G Yen.
\newblock Particle swarm optimization of deep neural networks architectures for
  image classification.
\newblock {\em Swarm and Evolutionary Computation}, 49:62--74, 2019.

\bibitem{wang2022dropout}
Ye-Qun Wang, Chun-Hua Chen, Jun Zhang, and Zhi-Hui Zhan.
\newblock Dropout topology-assisted bidirectional learning particle swarm
  optimization for neural architecture search.
\newblock In {\em Proceedings of the Genetic and Evolutionary Computation
  Conference Companion}, pages 93--96, 2022.

\bibitem{darwish2020survey}
Ashraf Darwish, Aboul~Ella Hassanien, and Swagatam Das.
\newblock A survey of swarm and evolutionary computing approaches for deep
  learning.
\newblock {\em Artificial intelligence review}, 53(3):1767--1812, 2020.

\bibitem{liu2018darts}
Hanxiao Liu, Karen Simonyan, and Yiming Yang.
\newblock Darts: Differentiable architecture search.
\newblock In {\em International Conference on Learning Representations}, 2018.

\bibitem{li2020random}
Liam Li and Ameet Talwalkar.
\newblock Random search and reproducibility for neural architecture search.
\newblock In {\em Uncertainty in artificial intelligence}, pages 367--377.
  PMLR, 2020.

\bibitem{zhang2020overcoming}
Miao Zhang, Huiqi Li, Shirui Pan, Xiaojun Chang, and Steven Su.
\newblock Overcoming multi-model forgetting in one-shot nas with diversity
  maximization.
\newblock In {\em Proceedings of the ieee/cvf conference on computer vision and
  pattern recognition}, pages 7809--7818, 2020.

\bibitem{kandasamy2018neural}
Kirthevasan Kandasamy, Willie Neiswanger, Jeff Schneider, Barnabas Poczos, and
  Eric~P Xing.
\newblock Neural architecture search with bayesian optimisation and optimal
  transport.
\newblock In {\em NeurIPS}, volume~31, 2018.

\bibitem{liu2018progressive}
Chenxi Liu, Barret Zoph, Maxim Neumann, Jonathon Shlens, Wei Hua, Li-Jia Li,
  Li~Fei-Fei, Alan Yuille, Jonathan Huang, and Kevin Murphy.
\newblock Progressive neural architecture search.
\newblock In {\em Proceedings of the European conference on computer vision
  (ECCV)}, pages 19--34, 2018.

\bibitem{simonyan2014very}
Karen Simonyan and Andrew Zisserman.
\newblock Very deep convolutional networks for large-scale image recognition.
\newblock In {\em International Conference on Learning Representations}, 2015.

\bibitem{zhu2020deep}
Yongchun Zhu, Fuzhen Zhuang, Jindong Wang, Guolin Ke, Jingwu Chen, Jiang Bian,
  Hui Xiong, and Qing He.
\newblock Deep subdomain adaptation network for image classification.
\newblock {\em IEEE Transactions on neural networks and learning systems},
  32(4):1713--1722, 2020.

\bibitem{shih2019real}
Kuan-Hung Shih, Ching-Te Chiu, Jiou-Ai Lin, and Yen-Yu Bu.
\newblock Real-time object detection with reduced region proposal network via
  multi-feature concatenation.
\newblock {\em IEEE Transactions on neural networks and learning systems},
  31(6):2164--2173, 2019.

\bibitem{zhang2020discriminant}
Dingwen Zhang, Junwei Han, Long Zhao, and Tao Zhao.
\newblock From discriminant to complete: Reinforcement searching-agent learning
  for weakly supervised object detection.
\newblock {\em IEEE Transactions on Neural Networks and Learning Systems},
  31(12):5549--5560, 2020.

\bibitem{wu2016google}
Yonghui Wu, Mike Schuster, Zhifeng Chen, Quoc~V Le, Mohammad Norouzi, Wolfgang
  Macherey, Maxim Krikun, Yuan Cao, Qin Gao, Klaus Macherey, et~al.
\newblock Google's neural machine translation system: Bridging the gap between
  human and machine translation.
\newblock {\em arXiv preprint arXiv:1609.08144}, 2016.

\bibitem{zhang2020neural}
Biao Zhang, Deyi Xiong, Jun Xie, and Jinsong Su.
\newblock Neural machine translation with gru-gated attention model.
\newblock {\em IEEE Transactions on neural networks and learning systems},
  31(11):4688--4698, 2020.

\bibitem{liang2018evolutionary}
Jason Liang, Elliot Meyerson, and Risto Miikkulainen.
\newblock Evolutionary architecture search for deep multitask networks.
\newblock In {\em Proceedings of the Genetic and Evolutionary Computation
  Conference}, pages 466--473, 2018.

\bibitem{kim2021auto}
Eunwoo Kim, Chanho Ahn, and Songhwai Oh.
\newblock Auto-virtualnet: Cost-adaptive dynamic architecture search for
  multi-task learning.
\newblock {\em Neurocomputing}, 442:116--124, 2021.

\bibitem{Asseman2021}
Alexis Asseman, Nicolas Antoine, and Ahmet~S. Ozcan.
\newblock {Accelerating Deep Neuroevolution on Distributed FPGAs for
  Reinforcement Learning Problems}.
\newblock {\em ACM Journal on Emerging Technologies in Computing Systems},
  17(2), 2021.

\bibitem{TFAgents}
Sergio Guadarrama, Anoop Korattikara, Oscar Ramirez, Pablo Castro, Ethan Holly,
  Sam Fishman, Ke~Wang, Ekaterina Gonina, Neal Wu, Efi Kokiopoulou, Luciano
  Sbaiz, Jamie Smith, Gábor Bartók, Jesse Berent, Chris Harris, Vincent
  Vanhoucke, and Eugene Brevdo.
\newblock {TF-Agents}: A library for reinforcement learning in tensorflow.
\newblock \url{https://github.com/tensorflow/agents}, 2018.
\newblock [Online; accessed 25-June-2019].

\bibitem{Gauci2019}
Jason Gauci, Edoardo Conti, Yitao Liang, Kittipat Virochsiri, Yuchen He,
  Zachary Kaden, Vivek Narayanan, Xiaohui Ye, Zhengxing Chen, and Scott
  Fujimoto.
\newblock {Horizon : Facebook ' s Open Source Applied Reinforcement Learning
  Platform}.
\newblock 2019.

\bibitem{parl}
Baidu.
\newblock A flexible, distributed and object-oriented programming reinforcement
  learning framework.
\newblock 2022.

\bibitem{Hoffman2020}
Matt Hoffman, Bobak Shahriari, John Aslanides, Gabriel Barth-Maron, Feryal
  Behbahani, Tamara Norman, Abbas Abdolmaleki, Albin Cassirer, Fan Yang, Kate
  Baumli, Sarah Henderson, Alex Novikov, Sergio~G{\'{o}}mez Colmenarejo, Serkan
  Cabi, Caglar Gulcehre, Tom~Le Paine, Andrew Cowie, Ziyu Wang, Bilal Piot, and
  Nando de~Freitas.
\newblock {Acme: A Research Framework for Distributed Reinforcement Learning}.
\newblock {\em arXiv preprint arXiv:2006.00979}, pages 1--33, 2020.

\bibitem{JMLR:v22:20-376}
Yasuhiro Fujita, Prabhat Nagarajan, Toshiki Kataoka, and Takahiro Ishikawa.
\newblock Chainerrl: A deep reinforcement learning library.
\newblock {\em Journal of Machine Learning Research}, 22(77):1--14, 2021.

\bibitem{tianshou}
Jiayi Weng, Huayu Chen, Dong Yan, Kaichao You, Alexis Duburcq, Minghao Zhang,
  Yi~Su, Hang Su, and Jun Zhu.
\newblock Tianshou: A highly modularized deep reinforcement learning library.
\newblock {\em Journal of Machine Learning Research}, 23(267):1--6, 2022.

\bibitem{bou2023torchrl}
Albert Bou, Matteo Bettini, Sebastian Dittert, Vikash Kumar, Shagun Sodhani,
  Xiaomeng Yang, Gianni~De Fabritiis, and Vincent Moens.
\newblock Torchrl: A data-driven decision-making library for pytorch, 2023.

\bibitem{sebastian2020memory}
Abu Sebastian, Manuel Le~Gallo, Riduan Khaddam-Aljameh, and Evangelos
  Eleftheriou.
\newblock Memory devices and applications for in-memory computing.
\newblock {\em Nature nanotechnology}, 15(7):529--544, 2020.

\bibitem{wang2020fpdeep}
Tianqi Wang, Tong Geng, Ang Li, Xi~Jin, and Martin Herbordt.
\newblock Fpdeep: Scalable acceleration of cnn training on deeply-pipelined
  fpga clusters.
\newblock {\em IEEE Transactions on Computers}, 69(8):1143--1158, 2020.

\bibitem{rao2022long}
Arjun Rao, Philipp Plank, Andreas Wild, and Wolfgang Maass.
\newblock A long short-term memory for ai applications in spike-based
  neuromorphic hardware.
\newblock {\em Nature Machine Intelligence}, 4(5):467--479, 2022.

\bibitem{hu2023asynchronous}
Zijian Hu, Xiaoguang Gao, Kaifang Wan, Qianglong Wang, and Yiwei Zhai.
\newblock Asynchronous curriculum experience replay: A deep reinforcement
  learning approach for uav autonomous motion control in unknown dynamic
  environments.
\newblock {\em IEEE Transactions on Vehicular Technology}, 72(11):13985--14001,
  2023.

\bibitem{liu2021regret}
Xu-Hui Liu, Zhenghai Xue, Jingcheng Pang, Shengyi Jiang, Feng Xu, and Yang Yu.
\newblock Regret minimization experience replay in off-policy reinforcement
  learning.
\newblock {\em Advances in neural information processing systems},
  34:17604--17615, 2021.

\bibitem{schwarke2023curiosity}
Clemens Schwarke, Victor Klemm, Matthijs van~der Boon, Marko Bjelonic, and
  Marco Hutter.
\newblock Curiosity-driven learning for joint locomotion and manipulation
  tasks.
\newblock In {\em 7th Annual Conference on Robot Learning}, 2023.

\bibitem{wu2022quality}
Shuang Wu, Jian Yao, Haobo Fu, Ye~Tian, Chao Qian, Yaodong Yang, Qiang Fu, and
  Yang Wei.
\newblock Quality-similar diversity via population based reinforcement
  learning.
\newblock In {\em International Conference on Learning Representations}, 2022.

\bibitem{bai2022training}
Yuntao Bai, Andy Jones, Kamal Ndousse, Amanda Askell, Anna Chen, Nova DasSarma,
  Dawn Drain, Stanislav Fort, Deep Ganguli, Tom Henighan, et~al.
\newblock Training a helpful and harmless assistant with reinforcement learning
  from human feedback.
\newblock {\em arXiv preprint arXiv:2204.05862}, 2022.

\bibitem{chu2023accelerating}
Kun Chu, Xufeng Zhao, Cornelius Weber, Mengdi Li, and Stefan Wermter.
\newblock Accelerating reinforcement learning of robotic manipulations via
  feedback from large language models.
\newblock {\em arXiv preprint arXiv:2311.02379}, 2023.

\bibitem{cao2024survey}
Yuji Cao, Huan Zhao, Yuheng Cheng, Ting Shu, Guolong Liu, Gaoqi Liang, Junhua
  Zhao, and Yun Li.
\newblock Survey on large language model-enhanced reinforcement learning:
  Concept, taxonomy, and methods.
\newblock {\em arXiv preprint arXiv:2404.00282}, 2024.

\end{thebibliography}

\end{document}